\documentclass[final,5p,times,twocolumn,authoryear]{elsarticle}

\usepackage[T1]{fontenc}
\usepackage[utf8]{inputenc}
\usepackage{amsmath,amssymb,bm}
\usepackage{graphicx}
\usepackage{booktabs}
\usepackage{multirow}
\usepackage{siunitx}
\usepackage{textcomp}
\usepackage[table]{xcolor}
\usepackage{url}
\usepackage{enumitem}
\usepackage{array}
\usepackage{placeins}
\usepackage{balance}
\usepackage{subcaption}
\usepackage[hidelinks,breaklinks=true]{hyperref}
\usepackage{xurl}

\DeclareSIUnit{\yr}{yr}
\DeclareSIUnit{\ha}{ha}
\graphicspath{{figures/minenetscd/}}

\newcommand{\colhead}[2][c]{\begin{tabular}[b]{@{}#1@{}}#2\end{tabular}}
\newcommand{\cellstack}[1]{\begin{tabular}[t]{@{}l@{}}#1\end{tabular}}

\newcommand{\TSeK}{\ensuremath{T_{\mathrm{SeK}}}}
\newcommand{\dA}{\ensuremath{\Delta A}}
\newcommand{\rprod}{\ensuremath{r}}
\newcommand{\rpart}{\ensuremath{r_{\mathrm{partial}}}}
\newcommand{\rinc}{\ensuremath{r_{\Delta A}}}

\begin{document}
\begin{frontmatter}
\title{MineNetTrace: A trajectory inference approach and Sentinel-2 benchmark for global mine monitoring}

\author[hzdr]{Weikang Yu\corref{cor1}}
\author[hzdr]{Vincent Nwazelibe}
\author[hzdr]{Sam Thiele}
\author[hzdr]{Richard Gloaguen}
\author[hzdr]{Pedram Ghamisi}
\cortext[cor1]{Corresponding author.}
\affiliation[hzdr]{organization={Helmholtz-Zentrum Dresden-Rossendorf}, city={Freiberg}, postcode={09599}, country={Germany}}

\begin{abstract}
Remote mine monitoring requires consistent and accurate mapping of mine extent from long time series of satellite images. This is difficult for conventional bitemporal change detection, which compares two images at a time. Here, we introduce TRACE-Net, a change detection approach that considers the full time-series trajectory of each pixel to infer when it was mined or converted to mine-related infrastructure. We demonstrate TRACE-Net on MineNetTrace, a new global benchmark dataset of 25 mines with 205 Sentinel-2 acquisitions, six-class masks of the first and last acquisition of each mine and harmonised mine production records. For eight test mines, the trajectory inference makes the maps temporally consistent without lowering the accuracy at the annotated dates: one-year false changes fall nearly fourteenfold and the pixel flip rate falls from 3.1\,\% to 0.6\,\%. The mapped one-year change is close to the ground truth. The same inference also raises the change accuracy and lowers the flip rate of other published semantic change detection architectures without retraining. On the EuroMineNet benchmark, TRACE-Net doubles the one-year change F1, and two masks per site train a network about as accurate as ten annual masks do. The mapped area increments are consistent with reported production across mines. This supports deriving annual maps of mine components with networks trained on sites annotated at only two dates.
\end{abstract}


\end{frontmatter}

\section{Introduction}
\label{sec:intro}

Mining causes large changes to the land surface and the environment. Global inventories estimate mining's direct footprint at roughly \SI{100000}{\kilo\metre\squared} \citep{maus2022update}, and further expansion threatens deforestation, biodiversity loss and associated pressure on protected areas and vulnerable ecosystems \citep{luckeneder2021surge,giljum2022pantropical}. These pressures are likely to grow as the energy transition and related high-tech industries increase demand for critical metals \citep{sonter2020renewable}.
Monitoring mines from space can help to provide much-needed data tracking their surface footprint and associated environmental impacts, assess whether planning and regulatory requirements are met, and help to increase the benefits of mining while limiting its harms \citep{ali2017mineral}. Remote-sensing based monitoring approaches are most useful if they consistently track changes in mine area and, importantly, the pre- and post-mining land use. Open pits, rock storage (waste-rock dumps and stockpiles), tailings facilities, process-water bodies and infrastructure pose different environmental risks and need to be managed differently.

Since 2015, Sentinel-2 has provided multispectral imagery at \SI{10}{\metre} resolution, every five days since the launch of its second satellite in 2017 \citep{drusch2012sentinel}. This resolution is sufficient to distinguish the main components of a mine, and the archive usually includes at least one cloud-free image per year. The archive therefore holds an annual, component-level record of the world's mines over the past decade.

Turning this archive into annual maps requires annotated training data. Semantic change detection (SCD) networks, for example, learn from annotated image pairs to predict the land-cover classes at two dates together with the change between them \citep{daudt2019multitask,yang2022asymmetric}. Annotating mine components, however, requires expert interpretation. Mining datasets, like most multitemporal remote-sensing datasets, therefore annotate one or two dates per site and leave the images in between unlabelled. We call this the \emph{endpoint-annotated setting}: dense semantic masks at the first and last acquisitions of a multi-year sequence, with no labels in between. We ask how every acquisition can be mapped in this setting, so that the change between any two dates can be derived, and how such maps can be evaluated without intermediate references. To be useful for monitoring, the maps must detect change between consecutive years, and the mapped areas must be accurate enough to track mine expansion.

Current approaches meet these requirements only in part. Bitemporal SCD networks ignore the images in between, and comparing the maps of different dates turns the errors of each map into false changes \citep{hafner2024continuous}. This is the central difficulty of annual monitoring: within one year, a mine changes a fraction of a per cent of its pixels, fewer than an accurate network misclassifies in a single map. On EuroMineNet, current methods accordingly captured long-term expansion more reliably than short-term change \citep{yu2025euromininet}. Satellite image time-series models use the whole sequence, but they either assign one class to each pixel for the entire sequence \citep{garnot2021panoptic,tarasiou2023vits} or require labels at every date during training. Temporal post-processing makes the maps of individual dates consistent, but it relies on fixed rules or on transition probabilities from dense reference data \citep{yang2021clcd,sullamenashe2019hierarchical,bogaert2022hidden}. None of these approaches uses the fact that each intermediate image lies between two known states.

Evaluation is a further challenge. SCD metrics are defined only at annotated dates, and the standard Separated Kappa (SeK) \citep{yang2022asymmetric}, under its common masking convention, discards most of the information on changes from background, which is the dominant change at mines. Finally, the best-established link between mine maps and a physical quantity relates total mine area to cumulative extraction at a single epoch \citep{werner2020global}, rather than the change of individual components.

Here, we use the fact that two annotated dates constrain both the classes and the changes between them. Mining mostly converts land outside the mine into one of the mine components, and this change is rarely reversed. A pixel can therefore be assumed to change at most once, and its unlabelled images become repeated observations of a single trajectory: one class before and one after a possible change, and the date of that change. Combining these observations helps correct errors that occur at single dates, which otherwise dominate the change between consecutive years. The network that provides the observations is trained on the endpoint masks of training sites, so that mapped sites need no annotation. On the test sites, false changes over one year fall nearly fourteenfold, while as many ground-truth changes are kept as in the per-frame maps. The same constraint allows the annotated dates to be used as a reference for evaluating the maps in between.

The main contributions of this paper are as follows:
\begin{enumerate}[leftmargin=*,itemsep=2pt]
\item \textbf{MineNetTrace: a global benchmark for the endpoint-annotated setting.} It comprises 25 mines on six continents observed yearly by Sentinel-2, with six-class masks at the first and last acquisition of each site, 155 unlabelled intermediate acquisitions, a fixed site split and harmonised production records as an independent reference.
\item \textbf{TRACE-Net: endpoint-informed trajectory inference.} A Siamese network trained on the endpoint pairs provides class and change evidence for every acquisition, and trajectory inference combines this evidence with a change prior into one trajectory per pixel with at most one change, so that the maps of all acquisitions are temporally consistent. The inference needs no further training and also improves other published SCD architectures.
\item \textbf{An evaluation protocol for endpoint-annotated sequences.} It comprises metrics for the unlabelled acquisitions, change metrics resolved by time span and a transition-space variant of SeK that keeps the changes from background, complemented by external validation on EuroMineNet, where two endpoint masks per site train a network about as accurate as annual masks.
\end{enumerate}

\section{Related work}
\label{sec:related}

\subsection{Earth observation of mining}
\label{sec:rw:mining}

Satellite mapping of mines has progressed from regional case studies to global inventories. Inventories of mining land use now comprise tens of thousands of polygons delineated with Sentinel-2 imagery \citep{maus2020global,maus2022update}, and one inventory distinguishes waste-rock dumps, pits, water ponds, tailings dams, heap-leach pads and processing infrastructure \citep{tang2023global}. They are based on expert delineation and represent one or a few epochs, so that they record where mines are rather than how they change from year to year. Their extents also depend on the delineation rules. The polygons of \citet{maus2022update} cover all land used by a mine, including enclosed patches of other land cover, but miss part of the mining area (producer's accuracy of \SI{78.9}{\percent}), whereas those of \citet{tang2023global} follow the individual mine features and give a smaller global footprint.

Learned methods have automated parts of this mapping. Convolutional networks have detected artisanal small-scale mining in Ghana and tailings dams in Brazil from Sentinel-2 imagery \citep{gallwey2020sentinel,balaniuk2020mining}, and MineNetCD extended learned mapping to bitemporal change detection at 100 mines worldwide, with high-resolution imagery and binary change labels \citep{minenetcd2024}. Time-series methods address time rather than composition: temporal segmentation with LandTrendr \citep{kennedy2010detecting} and continuous change detection and classification \citep{zhu2014continuous} recover the annual disturbance and reclamation of surface mines over decades \citep{xiao2020mapping,surfacemining2025}, and harmonic analysis of vegetation-index series dates the phases of vegetation disturbance around mines \citep{nwazelibe2026}. These methods map whether land is disturbed, not which component of a mine it belongs to. Learned maps thus resolve the components at one or two dates, whereas time-series maps resolve the years but not the components.

EuroMineNet made time an explicit dimension of learned mine mapping. It provides expert-verified annual mining footprints for 133 European sites from 2015 to 2024 and benchmarks both annual footprint mapping and change detection between years \citep{yu2025euromininet}. The only mining benchmark with annual masks thus covers one continent and binary footprints, and labelling every year at component level would multiply the annotation effort required for a global dataset.

Mine maps are usually validated against reference delineations of the same kind, which shows whether a map agrees with an interpreter but not whether it measures the activity of the mine. At global scale, cumulative production explains much of the variation in total mine area across 129 mines, measured on whole-mine polygons at a single epoch \citep{werner2020global}. Whether the change of individual components between two dates tracks the material moved, and whether learned maps preserve such a relationship, has to our knowledge not been tested. MineNetTrace addresses these gaps with component-level masks at the endpoints of 25 mines on six continents, the annual acquisitions between them and harmonised production records, which provide an independent reference for the mapped area increments.

\subsection{Semantic change detection and its metrics}
\label{sec:rw:scd}

Deep change detection began with the binary change between two dates: fully convolutional Siamese networks compare the features of two co-registered images \citep{daudt2018fully}, and transformer-based detectors model the context between the dates \citep{chen2022bit}; \citet{cheng2024change} review the methods of the last decade. Semantic change detection extends the question from where a scene changed to what it changed from and what it changed into. Multitask networks jointly predict land cover at both dates and a change mask \citep{daudt2019multitask}, and the SECOND and Hi-UCD benchmarks established the task for aerial imagery of urban scenes \citep{yang2022asymmetric,tian2022large}.

Subsequent SCD architectures differ mainly in how the two dates interact and how the semantic and change outputs are tied together. ChangeMask separates semantic segmentation from change detection within a temporal-symmetric design \citep{zheng2022changemask}, BiSRNet adds semantic reasoning within and across the dates \citep{ding2022bisrnet}, and TED decodes the two semantic maps and the change in three branches from a shared encoder \citep{ding2024scannet}. Later networks enforce consistency between the semantic and change outputs (CdSC, SSCD-L; \citealp{wang2024cdsc,jiang2025sscdl}), guide the semantic branches by change information, object boundaries or change priors (EGMSNet, BGSNet, CPGNet; \citealp{zuo2025egmsnet,long2025bgsnet,long2025cpgnet}), reason over hierarchical semantic graphs (HGINet; \citealp{long2024hginet}), fuse the two dates late so that pretrained backbones can be used unchanged (LSAFNet; \citealp{zhou2025lsafnet}), adopt state-space models (ChangeMamba, GSTM-SCD; \citealp{chen2024changemamba,gstm2025}), combine foreground and background guidance (FoBa; \citealp{foba2025}) or treat the image pair as a short video (Change3D; \citealp{zhu2025change3d}). Together with two of the multitask strategies of \citet{daudt2019multitask} (HRSCD), these architectures form the 16 baselines of this study. All of them are trained on pairs of annotated dates and predict the change between the two dates they are given.

Longer sequences have been addressed in two ways. Models for satellite image time series, from recurrent encoders \citep{russwurm2018multi} to temporal self-attention \citep{garnot2020pse,garnot2021panoptic} and time-series transformers \citep{tarasiou2023vits}, segment a scene from its whole sequence, typically into one map per sequence, and are trained with a label for every training sequence. Multitemporal change benchmarks such as SpaceNet~7 \citep{vanetten2021spacenet} and DynamicEarthNet \citep{toker2022dynamicearthnet} instead label every date of a longer sequence, at much greater annotation cost. Where only some dates are labelled, each date is usually mapped separately and the maps are compared afterwards, which accumulates the errors of the individual maps \citep{hafner2024continuous}. The related benchmarks therefore follow one of two supervision patterns (Table~\ref{tab:datasets}): SECOND, Hi-UCD and MineNetCD label every acquisition of only two or three dates per site, leaving no intermediate dates to infer, whereas SpaceNet~7, DynamicEarthNet and EuroMineNet label every date. None of them includes an independent physical reference.

\begin{table*}[tp]
\centering
\caption{Related multitemporal benchmarks, as described in their publications. GSD: ground sample distance; Dates: acquisitions per site or area; labelled: acquisitions with dense masks.}
\label{tab:datasets}
\footnotesize
\setlength{\tabcolsep}{3pt}
\begin{tabular*}{\linewidth}{@{\extracolsep{\fill}}lllllll@{}}
\toprule
Benchmark & Domain & Sensor, GSD & Sites & Dates (labelled) & Semantic classes & External reference \\
\midrule
\cellstack{SECOND\\\citep{yang2022asymmetric}} & urban, China & aerial, \SIrange{0.5}{3}{\metre} & 4\,662 tile pairs & 2 (2) & 6 land-cover & none \\\addlinespace[2pt]
\cellstack{Hi-UCD\\\citep{tian2022large}} & urban, Tallinn & aerial, \SI{0.1}{\metre} & 1 city & 3 (3) & 9 land-cover & none \\\addlinespace[2pt]
\cellstack{SpaceNet~7\\\citep{vanetten2021spacenet}} & urban, global & Planet, \SI{4}{\metre} & $>$100 areas & 24 monthly (24) & buildings & none \\\addlinespace[2pt]
\cellstack{DynamicEarthNet\\\citep{toker2022dynamicearthnet}} & land cover, global & Planet, \SI{3}{\metre} & 75 areas & \cellstack{daily, 24 months\\(24 monthly)} & 7 land-cover & none \\\addlinespace[2pt]
\cellstack{MineNetCD\\\citep{minenetcd2024}} & mining, global & \cellstack{high-resolution\\optical} & 100 sites & 2 (2) & binary change & none \\\addlinespace[2pt]
\cellstack{EuroMineNet\\\citep{yu2025euromininet}} & mining, EU & Sentinel-2, \SI{10}{\metre} & 133 sites & \cellstack{annual 2015--2024\\(all)} & footprint (binary) & none \\\addlinespace[2pt]
\cellstack{\textbf{MineNetTrace}\\(this paper)} & mining, global & Sentinel-2, \SI{10}{\metre} & \cellstack{25 sites,\\6 continents} & \cellstack{6--11 annual,\\2015--2024 (2)} & 6 mine components & annual production \\
\bottomrule
\end{tabular*}
\end{table*}

Evaluation follows the same pattern. SCD benchmarks report the mean intersection over union (mIoU) of the unchanged and the changed pixels, a semantic change F-score and SeK, introduced with SECOND, which removes the no-change class before computing agreement and scales the result by the intersection over union of the changed pixels \citep{yang2022asymmetric}. Under the common masking convention, SeK also assigns every changed pixel whose source class is background to the no-change index; where most change starts from background, as at mines, this discards most of the information that the metric is meant to summarise. Every SCD metric is also defined only at annotated dates. This study keeps bitemporal SCD networks as the source of evidence at each date but derives the maps of all acquisitions from one trajectory per pixel, and it complements the standard metrics with a transition-space variant of SeK and with metrics for the unlabelled acquisitions.

\subsection{Temporal consistency of land-cover time series}
\label{sec:rw:temporal}

Annual land-cover maps are commonly produced by classifying each acquisition or annual composite separately; Dynamic World, for example, classifies every Sentinel-2 acquisition independently \citep{brown2022dynamic}. Errors that vary between dates then appear as changes when the maps of two dates are compared \citep{zhu2017change}, and where real change is rare, as between consecutive years, these false changes can outnumber the real ones. For mines, EuroMineNet made temporal consistency an explicit evaluation criterion: its Change-Aware Temporal IoU scores the agreement between the maps of two years where the annual reference labels do not change, and its benchmark of current deep learning models found that long-term change was captured reliably, whereas short-term dynamics remained difficult to detect \citep{yu2025euromininet}. Consistent annual maps therefore require that the evidence of different dates be combined.

Post-classification approaches enforce consistency on the sequence of mapped classes. Rule-based temporal filters replace a class observed in a single year that lies between years of another class, as in MapBiomas \citep{souza2020reconstructing} and the annual land-cover dataset of China \citep{yang2021clcd}. Hidden Markov models treat the mapped classes as noisy observations of a hidden class sequence and so separate real change from classification noise \citep{abercrombie2016improving,bogaert2022hidden}, and \citet{hafner2024continuous} integrate the segmentation and change outputs of a network with a Markov network into a consistent building time series. These methods score a class sequence by its transitions between consecutive dates, with transition probabilities that are assumed or estimated from dense reference data, and most of them permit any number of changes along the sequence.

Temporal segmentation models the trajectory of a pixel rather than its class sequence: LandTrendr \citep{kennedy2010detecting}, BFAST \citep{verbesselt2010bfast} and continuous change detection and classification \citep{zhu2014continuous} fit the time series of a pixel with a few segments separated by breaks, so that all dates within a segment share one class and the change between any two dates follows from the breaks. Seasonal Harmonic Anomaly Break Analysis (SHABA; \citealp{nwazelibe2026}) instead locates multiple breaks in the deviation of a vegetation-index series from its seasonal harmonics. These methods segment a continuous spectral variable, and those that model seasonality require several observations per year.

The endpoint-annotated setting provides what these approaches assume or learn from dense reference data. The annotated endpoints of the training sites give the classes before and after change and the probability that a pixel changes over a period. Because a mining pixel rarely changes more than once within a decade, it can be assumed to change at most once, so that only its classes and the date of the change remain to be inferred. TRACE-Net combines the probabilistic and the trajectory-based approaches. Like temporal segmentation, it describes each pixel by a single trajectory, from which the maps of all dates and the change between any two dates follow; like the Markov models, it combines learned evidence of classes and of change with a prior. Its prior, however, is learned from the endpoint annotations alone and placed on the whole trajectory rather than on the transitions between consecutive dates.

\section{The MineNetTrace benchmark}
\label{sec:data}

\subsection{Sites, imagery and annotations}
\label{sec:data:sites}

MineNetTrace contains 25 mines observed by Sentinel-2 between 2015 and 2024 (Fig.~\ref{fig:design}; site inventory in Table~\ref{tab:sites}). They span six continents, from \ang{41.5}~S to \ang{67.7}~N, environments from boreal to tropical, and seven commodity groups: copper, gold, iron, molybdenum, diamond, zinc--lead and polymetallic copper--zinc. Twenty-one are open-pit operations; the other four are underground mines, and the scene of one of them, Mount Isa, also contains former open pits. Each site is a single co-registered scene covering its operational footprint (\SI{4866}{\kilo\metre\squared} in total), with 6 to 11 acquisitions (median 8), 205 in total. The imagery is Sentinel-2 Level-1C top-of-atmosphere reflectance in nine bands (B2, B3, B4, B5, B6, B7, B8, B11 and B12) at \SI{10}{\metre}, with a per-pixel validity mask. Acquisitions were chosen to be cloud-free over the footprint, at approximately annual intervals and, where possible, from the same season; 174 of the 180 intervals between consecutive acquisitions last between 0.5 and 1.5 years (Fig.~\ref{fig:design}d). The residual misregistration between consecutive acquisitions, estimated by phase correlation \citep{guizarsicairos2008efficient}, has a median of 0.2 pixels; one pair in ten, mostly before 2022, is offset by about one pixel.

The first and last acquisitions of every site carry dense masks for six classes: \emph{background}; \emph{pit}, the open-pit excavations; \emph{rock storage}, the waste-rock dumps and stockpiles; \emph{tailings}; \emph{infrastructure}, including processing plants, camps and other built facilities; and \emph{water}, including process-water reservoirs, pit lakes and dams. A remote-sensing interpreter annotated each endpoint, using very-high-resolution imagery from Google Earth as a reference for class boundaries, and another remote-sensing expert verified the masks. Class prevalence is highly imbalanced: pooled over the 50 masks, background occupies \SI{85.92}{\percent} of the pixels and water only \SI{0.37}{\percent} (Fig.~\ref{fig:design}c). We call the union of the five mining classes the \emph{mining area}; at the final endpoint it ranges from \SI{2.1}{\kilo\metre\squared} at Black Mountain to \SI{199}{\kilo\metre\squared} at Chuquicamata. The 155 acquisitions between the endpoints are released without masks.

\begin{figure*}[tp]
\centering
\input{figures/minenetscd/fig01_study_design_nb_panels.tex}
\caption{Overview of MineNetTrace. (a) Locations of the 25 sites, coloured by principal commodity; marker shape gives the split (basemap: Esri World Light Gray Base; Esri, HERE, Garmin, OpenStreetMap contributors). (b) Acquisition timeline per site: annotated endpoints (squares, coloured by split) and unlabelled acquisitions (circles). (c) Class prevalence over the 50 annotated masks. (d) Intervals between consecutive acquisitions.}
\label{fig:design}
\end{figure*}

\begin{table*}[tp]
\centering
\caption{The 25 sites of MineNetTrace. Acq.: acquisitions; Area: scene extent; Split: role in the fixed split; last column: production indicators that meet the coverage rule (P ore processed, M ore mined, T total material mined, C main-commodity output; en dash: none). UG: underground operation.}
\label{tab:sites}
\footnotesize
\setlength{\tabcolsep}{4pt}
\begin{tabular*}{\linewidth}{@{\extracolsep{\fill}}llllrrrll@{}}
\toprule
Site & Country & Commodity & Lat., Lon. & Area (\si{\kilo\metre\squared}) & Acq. & Period & Split & Production \\
\midrule
Aitik          & Sweden       & Cu      & \ang{67.1}~N, \ang{20.9}~E   &  209 &  6 & 2019--2024 & train & P \\
Awonsu (Ahafo South) & Ghana  & Au      & \ang{7.0}~N, \ang{2.4}~W     &  183 & 11 & 2015--2024 & train & P, M, T, C \\
Black Mountain (UG) & South Africa & Zn--Pb  & \ang{29.3}~S, \ang{18.8}~E   &   26 &  8 & 2017--2024 & train & -- \\
Boddington     & Australia    & Au      & \ang{32.7}~S, \ang{116.4}~E  &  201 &  8 & 2017--2024 & test  & P, M, T, C \\
Cannington (UG) & Australia   & Zn--Pb  & \ang{21.8}~S, \ang{140.9}~E  &   61 &  7 & 2018--2024 & test  & P, M, C \\
Chuquicamata   & Chile        & Cu      & \ang{22.3}~S, \ang{68.8}~W   & 1290 &  8 & 2018--2024 & train & C \\
Climax         & USA          & Mo      & \ang{39.4}~N, \ang{106.2}~W  &  274 &  7 & 2018--2024 & test  & C \\
Constancia     & Peru         & Cu      & \ang{14.5}~S, \ang{71.8}~W   &   99 &  8 & 2017--2024 & test  & P, M, C \\
Damtshaa       & Botswana     & Diamond & \ang{21.3}~S, \ang{25.5}~E   &   16 &  9 & 2016--2024 & train & C \\
Erdenet        & Mongolia     & Cu      & \ang{49.1}~N, \ang{104.1}~E  &  239 &  8 & 2017--2024 & train & P, M \\
Essakane       & Burkina Faso & Au      & \ang{14.4}~N, \ang{0.1}~E    &  180 &  9 & 2016--2024 & test  & P, M, T, C \\
Jwaneng        & Botswana     & Diamond & \ang{24.5}~S, \ang{24.7}~E   &  126 &  9 & 2016--2024 & val   & C \\
Kansanshi      & Zambia       & Cu      & \ang{12.1}~S, \ang{26.4}~E   &  401 &  9 & 2016--2024 & train & P, M, T, C \\
Kevitsa        & Finland      & Cu      & \ang{67.7}~N, \ang{27.0}~E   &  118 & 10 & 2015--2024 & test  & P, M, T, C \\
Mission        & USA          & Cu      & \ang{32.0}~N, \ang{111.1}~W  &  152 &  8 & 2017--2024 & train & C \\
Mount Isa (UG) & Australia    & Zn--Pb  & \ang{20.7}~S, \ang{139.4}~E  &  119 &  8 & 2017--2024 & val   & C \\
Mt Rawdon      & Australia    & Au      & \ang{25.3}~S, \ang{151.8}~E  &   17 &  9 & 2016--2024 & train & C \\
Neves-Corvo (UG) & Portugal   & Cu--Zn  & \ang{37.6}~N, \ang{8.0}~W    &   39 &  8 & 2017--2024 & val   & P, M, C \\
Olimpiada      & Russia       & Au      & \ang{59.9}~N, \ang{92.9}~E   &  184 &  8 & 2016--2023 & train & P, M, T, C \\
Ro\c{s}ia Poieni & Romania    & Cu      & \ang{46.3}~N, \ang{23.2}~E   &   54 &  9 & 2017--2024 & train & -- \\
Savage River   & Australia    & Fe      & \ang{41.5}~S, \ang{145.2}~E  &  126 &  6 & 2018--2024 & test  & C \\
Sierra Gorda   & Chile        & Cu      & \ang{22.9}~S, \ang{69.4}~W   &  168 &  9 & 2016--2024 & train & -- \\
Sossego        & Brazil       & Cu      & \ang{6.4}~S, \ang{50.1}~W    &   99 &  9 & 2016--2024 & test  & M, C \\
Tilden         & USA          & Fe      & \ang{46.4}~N, \ang{87.6}~W   &  381 &  6 & 2019--2024 & train & M \\
Zijin Tongshan & China        & Cu      & \ang{50.2}~N, \ang{125.8}~E  &  102 &  8 & 2017--2024 & val   & C \\
\bottomrule
\end{tabular*}
\end{table*}

\begin{figure*}[tp]
\centering
\input{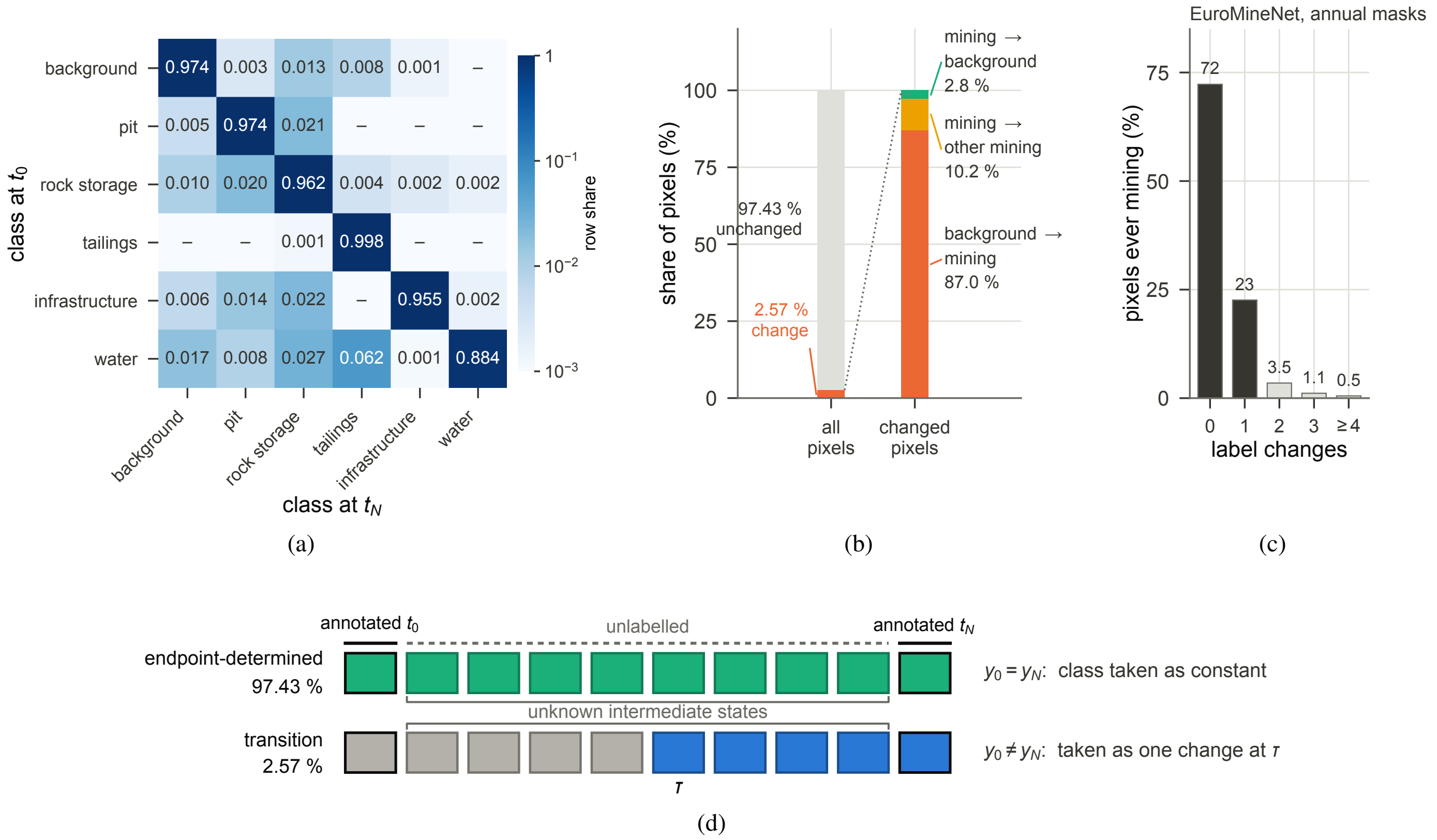}
\caption{The endpoint constraint. (a) Endpoint transition matrix over the 25 sites (rows sum to one; logarithmic colour scale; dash: below 0.001). (b) Unchanged and changed pixels, and the types of change. (c) Pixels labelled as mining in at least one year of the annual EuroMineNet masks (133 sites), by the number of label changes from 2015 to 2024. (d) Assumed trajectories of an endpoint-determined and a transition pixel (outlined: annotated; hatched: unlabelled; $\tau$: change date).}
\label{fig:bracket}
\end{figure*}

\subsection{The endpoint constraint}
\label{sec:data:constraint}

The endpoint pair has a structure on which the rest of the paper relies (Fig.~\ref{fig:bracket}). The two masks agree for \SI{97.43}{\percent} of the pixels. Of the \SI{2.57}{\percent} that change over a mean endpoint period of \SI{7.35}{\yr}, \SI{87.0}{\percent} change from background into a mining class, \SI{10.2}{\percent} from one mining class into another and \SI{2.8}{\percent} from a mining class back to background, reflecting reclamation, natural succession and backfilling. Change thus runs almost entirely in one direction, from background into the mining classes. Per-site change fractions range from \SI{0.25}{\percent} to \SI{12.3}{\percent}.

For a pixel whose two endpoint labels agree, we assume that its class is the same at every intermediate date; for a pixel whose labels differ, we assume that it changes once, at an unknown date, from the from-class to the to-class. We call this the \emph{single-change assumption}. Because change runs almost entirely in one direction, these are the most likely trajectories, and the annual masks of EuroMineNet show that repeated changes are rare (Fig.~\ref{fig:bracket}c). We call pixels whose endpoint labels agree \emph{endpoint-determined} and the others \emph{transition pixels}; on the eight test sites, endpoint-determined pixels make up \SI{98.1}{\percent} of the annotated area. For evaluation, the endpoint masks thus fix the label of every endpoint-determined pixel at every intermediate date, and only the date of change of a transition pixel remains open (Fig.~\ref{fig:bracket}d). We call the two annotated states, which together constrain the label of every pixel at every intermediate date, the \emph{endpoint constraint}. It underlies both the trajectory inference and the endpoint-derived evaluation of this paper.

\subsection{Data split}
\label{sec:data:split}

The fixed split contains 13 training, 4 validation and 8 test sites (Fig.~\ref{fig:design}a). Training contains \SI{68}{\percent} of both valid pixels and changed area, and every commodity group appears among the validation and test sites.

\subsection{Production records as an external reference}
\label{sec:data:production}
\label{sec:prod}

\begin{figure*}[tp]
\centering
\input{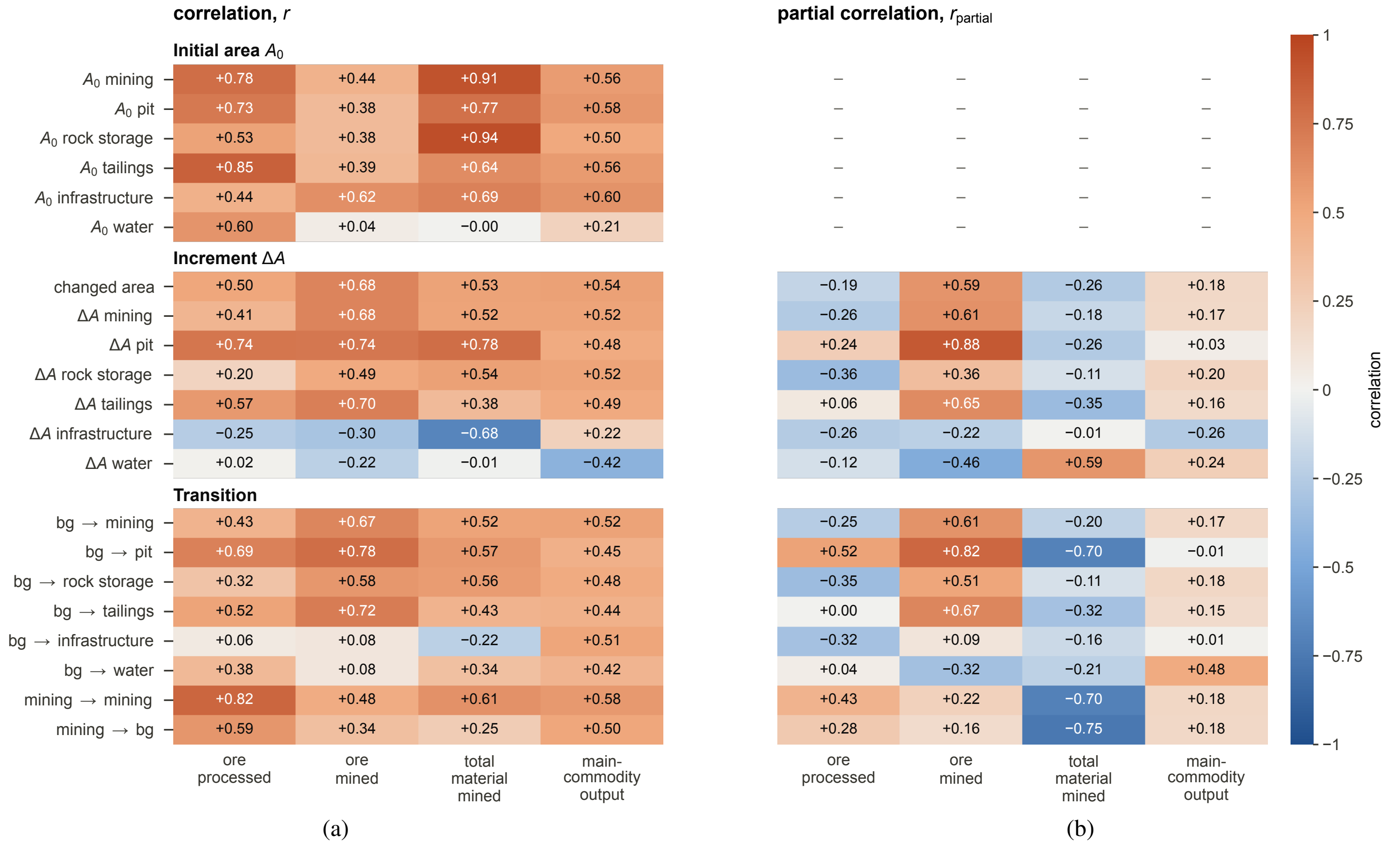}
\caption{Correlation of the map-derived variables with production, from the annotated masks of the 25 sites. (a) Cross-mine correlation \rprod{} with each production indicator. (b) Partial correlation \rpart{}, controlling for initial mining area (dash: not computed). bg: background.}
\label{fig:prelim}
\end{figure*}

For each site, we compiled operator-reported annual production from company technical and sustainability reports into a harmonised table with per-value quality flags. Four production indicators are defined: \emph{ore processed}, the tonnage fed to the mill; \emph{ore mined}, the tonnage of ore extracted; \emph{total material mined}, the tonnage of ore plus waste rock moved; and \emph{main-commodity output}, the quantity of the principal product. Because this quantity is reported in commodity-specific units, it is standardised within commodity groups of at least three sites for the cross-mine analysis (12 gold and copper mines). An indicator is used for an acquisition interval only when the reporting years cover at least \SI{85}{\percent} of that interval, weighted by month; we call such intervals \emph{qualifying}. The complete endpoint period qualifies at 11 sites for ore processed, 12 for ore mined and 6 for total material mined (Table~\ref{tab:sites}). Production is used only for evaluation.

Before fitting any model, we use the annotations of all 25 sites to ask which mapped quantities reflect the amount of material moved. From the endpoint masks, we derive three families of map-derived variables: the \emph{initial area} $A_0$ of each class and of the mining area; its \emph{increment} $\dA$ between the endpoints, together with the total changed area; and the \emph{transitions}, the areas that changed from one class to another. Increments and transitions are divided by the duration of the endpoint period. For each pair of variable and production indicator, every qualifying mine contributes its map-derived value and its mean production rate over the same period, and \rprod{} denotes their Pearson correlation across mines. Mine size is an important confounder: the initial mining area alone correlates strongly with ore processed ($\rprod=+0.78$). We therefore also report the partial correlation \rpart{} after linearly removing the initial mining area from both quantities, which asks whether a mapped quantity adds information beyond mine size. With 144 correlations in total (Fig.~\ref{fig:prelim}a,b), individual correlations are not marked significant; for those on which the paper relies, Table~\ref{tab:prodci} gives 95\,\% intervals (Fisher $z$), Holm-adjusted $p$-values over all 144 correlations and Spearman rank correlations $\rho$.

The associations differ strongly between indicators and between classes. For ore processed and total material mined, the strongest correlates are initial areas, and the correlations of the change terms weaken or become negative after size adjustment, so that these associations mainly reflect mine size. Ore mined is the only indicator whose change terms remain strong after adjustment. The pit increment tracks it most closely once mine size is accounted for ($\rpart=+0.88$, 95\,\% interval $+0.60$ to $+0.97$); this association is the only one that remains significant after Holm correction (adjusted $p=0.045$). The mining-area increment also tracks ore mined across mines ($\rprod=+0.68$), and most of this association persists beyond mine size ($\rpart=+0.61$), although neither value remains significant after the correction. Tailings follow the mining area, rock storage is weaker, and neither the infrastructure nor the water increment shows a clear positive raw correlation with any tonnage indicator. Without Tilden (open marker in Fig.~\ref{fig:prelim_scatter}), a large mine whose mining area changed little while production continued, the size-adjusted mining-area and tailings correlations become weak. Main-commodity output, which combines throughput with grade and recovery, correlates less strongly with the increments than ore mined.

The pit may stand out because a map records area but not height. Pit walls must keep a stable slope, so that a pit that deepens also widens, and extraction appears as lateral expansion, whereas tailings facilities are raised in successive lifts behind their dams and waste-rock dumps are built up in benches, so that much of the added material increases height rather than area.

We therefore use ore mined as the main indicator for validation and focus on two responses: the total mining area and the pit area, the latter corrected for initial mine size. Underground mines, which extract ore without a pit, enter the pit correlations with a pit increment of zero; their surface footprint grows through rock storage, tailings and infrastructure, which the mining area includes.

\section{TRACE-Net: trajectory inference from anchored change evidence}
\label{sec:method}

\subsection{Problem statement and overview}
\label{sec:method:overview}

Consider a site with Sentinel-2 acquisitions $x_0,\ldots,x_N$ at times $t_0<\cdots<t_N$. A training site also has dense masks $y_0$ and $y_N$ at the endpoints; a site to be mapped has no masks, so its image series is the only input. The goal is to predict a semantic map $\hat{y}_t$ and class areas $A_c(t)$ for every acquisition, where $c$ indexes the six classes.

\begin{figure*}[tp]
\centering
\includegraphics[width=\textwidth]{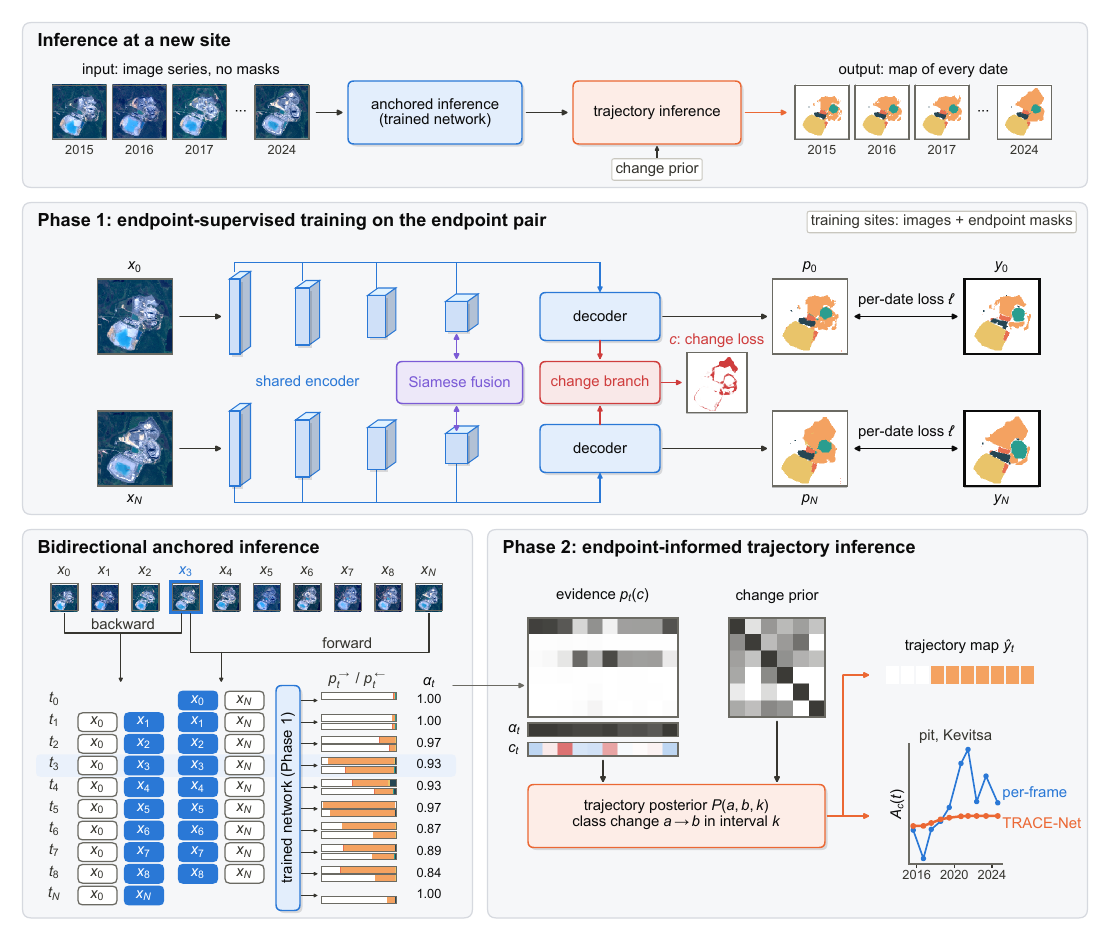}
\caption{Overview of TRACE-Net. Top: inference at a new site, whose image series is the only input; masks are used only to train Phase~1 and to estimate the change prior. Below: Phase~1 on the endpoint pair of a training site, bidirectional anchored inference and endpoint-informed trajectory inference (Kevitsa; pixel of Fig.~\ref{fig:phase2target}). $p_t$: class probabilities; $c$, $c_t$: change logits; $\alpha_t$: bidirectional agreement; $\hat{y}_t$: trajectory map; $\ell$: per-date loss.}
\label{fig:pipeline}
\end{figure*}

TRACE-Net has two phases (Fig.~\ref{fig:pipeline}). Phase~1 trains a Siamese SCD network on the endpoint pairs of the training sites: the two dates share a pretrained encoder, exchange semantic context through Siamese fusion, and are compared by a change branch. At inference, bidirectional anchored inference applies the trained network to every acquisition paired with the first and with the last image of the series and provides the class probabilities of the acquisition, the agreement of the two predictions and a change logit. Phase~2, endpoint-informed trajectory inference (hereafter trajectory inference), combines this evidence over all acquisitions of a pixel, weighting each acquisition by its bidirectional agreement. It assumes that the pixel changes at most once and uses a change prior estimated from the annotations of the training sites, and so maps every acquisition of a site without annotation. Phase~2 requires no training, only the class probabilities of a network and its change logits where it has a change output, so that it also applies to other segmentation and SCD architectures.

\subsection{Phase 1: endpoint-supervised Siamese network with fusion and change branch}
\label{sec:method:phase1}

\paragraph{Input and backbone} The input consists of the nine Sentinel-2 bands, standardised with robust statistics (per-band median and interquartile range) fitted on the training sites and multiplied by the validity mask. The segmentation network $f_\theta$ applied to each date is SegFormer-B2 \citep{xie2021segformer}, an ImageNet-initialised encoder--decoder \citep{iakubovskii2019smp}, whose first convolution is extended to nine channels by repeating the pretrained RGB kernels with rescaled weights; three other backbones are compared in the ablations.

\paragraph{Siamese fusion} The two dates of a pair pass through a shared encoder. At its deepest stage, whose features describe the semantic context of each pixel, a fusion module lets each date draw on the features of the other. With $u$ and $u'$ the deepest feature maps of the two dates,
\begin{equation}
\label{eq:fusion}
u \leftarrow u + W_2\,\mathrm{ReLU}\!\left(\mathrm{GN}\!\left(W_1\,\big[\,u;\ u';\ |u-u'|\,\big]\right)\right),
\end{equation}
and symmetrically for $u'$, where $[\cdot;\cdot]$ concatenates channels, $W_1$ and $W_2$ are pixel-wise linear maps and GN is group normalisation. The concatenation passes the features of both dates and their absolute difference marks where the dates disagree, as in Siamese change-detection networks \citep{daudt2018fully}. $W_2$ starts at zero, so that training begins from the pretrained single-date network and learns how much each date draws on the other. The shallower stages, which carry the spatial detail of each date, are not fused; a decoder and a semantic head map the features of each date to class logits.

\paragraph{Change branch} A change branch compares the decoded features of the two dates and returns at every pixel a logit $c$ that the class differs between them. It is trained on the endpoint pair against $y_0\neq y_N$, with the changed pixels weighted by the inverse of their share $\phi$ in the training annotations. This weighting cancels the prior odds of change, so that $c$ estimates the log-likelihood ratio of change and can be combined in Phase~2 with the change prior estimated from the annotations. A likelihood ratio, unlike a probability of change, does not depend on how often change occurs over the period of a pair, so that the logit learned on the endpoint pairs also applies to the shorter pairs of anchored inference.

\paragraph{Training objective} The network is trained on the endpoint pair $(x_0, x_N)$ of each training site, with class logits $z_0$ and $z_N$ for the two dates. The Phase~1 objective is
\begin{equation}
\label{eq:loss1}
\begin{split}
\mathcal{L}_1 ={}& \ell(z_0, y_0) + \ell(z_N, y_N) \\
&+ \lambda_{\mathrm{c}}\,\mathrm{BCE}_{1/\phi}\big(c_{0N},\, \mathbb{1}[y_0\neq y_N]\big)
+ \lambda_{\mathrm{a}}\,\mathcal{L}_{\mathrm{area}} ,
\end{split}
\end{equation}
with $c_{0N}$ the change logit of the endpoint pair, $\ell(z,y)=\mathrm{CE}(z,y)+\lambda_{\mathrm d}\mathrm{Dice}(z,y)$ and $\lambda_{\mathrm c}=1$. $\mathrm{CE}$ is a label-smoothed cross-entropy with class weights proportional to $p_c^{-1/2}$, where $p_c$ is the training prevalence of class $c$, and the soft Dice term covers the five mining classes ($\lambda_{\mathrm d}=0.5$). The area-consistency term $\mathcal{L}_{\mathrm{area}}$ is a Huber loss between predicted and annotated class-area increments, with areas computed from summed class probabilities ($\lambda_{\mathrm a}=0.2$).

Training tiles of $256\times 256$ pixels are drawn with replacement at weights of 8:2:1 for change tiles (at least \SI{0.2}{\percent} of valid pixels changed), mining tiles and background tiles, so that the rare change tiles are seen often while the static tiles are kept. Geometric augmentation is applied identically to both dates of a pair, and a radiometric augmentation (per-channel gain and offset with a standard deviation of 0.10) is drawn once per pair, so that augmentation never introduces spurious temporal change. We use AdamW with learning rates of $10^{-4}$ for the decoder and heads and $10^{-5}$ for the encoder, a one-cycle schedule and 80 epochs of 200 steps at a batch size of 16, and select the checkpoint by the change metric \TSeK{} (Section~\ref{sec:impl:protocol}) averaged over the four validation sites.

\paragraph{Bidirectional anchored inference} The trained network maps an intermediate acquisition $x_t$ by pairing it with an endpoint image, which we call \emph{anchored inference}: in its forward pair $(x_t, x_N)$ and in its backward pair $(x_0, x_t)$. Because each pair contains an endpoint, its change logit indicates whether the pixel changes between $x_t$ and that endpoint, that is, after or before $t$. TRACE-Net predicts every intermediate acquisition in both pairs, which we call \emph{bidirectional anchored inference}, and the endpoints in their pair $(x_0, x_N)$. The forward and backward predictions concern the same image in different temporal contexts, and where they disagree the evidence of the acquisition is uncertain. The class evidence of acquisition $t$ is therefore the mean $p_t$ of its forward and backward class probabilities, $p_t^{\rightarrow}$ and $p_t^{\leftarrow}$, together with their \emph{bidirectional agreement}
\begin{equation}
\label{eq:agree}
\alpha_t = 1 - \mathrm{JS}\big(p_t^{\rightarrow},\, p_t^{\leftarrow}\big) / \log 2 ,
\end{equation}
where $\mathrm{JS}$ is the Jensen--Shannon divergence, so that $\alpha_t$ is 1 where the two predictions agree and 0 where they predict different classes with certainty ($\alpha_t = 1$ at the endpoints). Trajectory inference weights the evidence $p_t$ of each acquisition by $\alpha_t$ (Eq.~\eqref{eq:switch}), so that the two predictions enter as an \emph{agreement-weighted mean}; their mean, which gives every acquisition full weight, is the special case $\alpha_t = 1$. The maps obtained from $p_t$ alone are the \emph{per-frame maps}. Anchored inference also returns a change logit $c_t$ between $x_t$ and the farther endpoint, from its forward pair when $2t \leq N$ and from its backward pair otherwise. Whole-scene maps average the softmax probabilities and change logits of overlapping tiles (64-pixel overlap).

\subsection{Phase 2: endpoint-informed trajectory inference}
\label{sec:method:phase2}

\begin{figure*}[tp]
\centering
\input{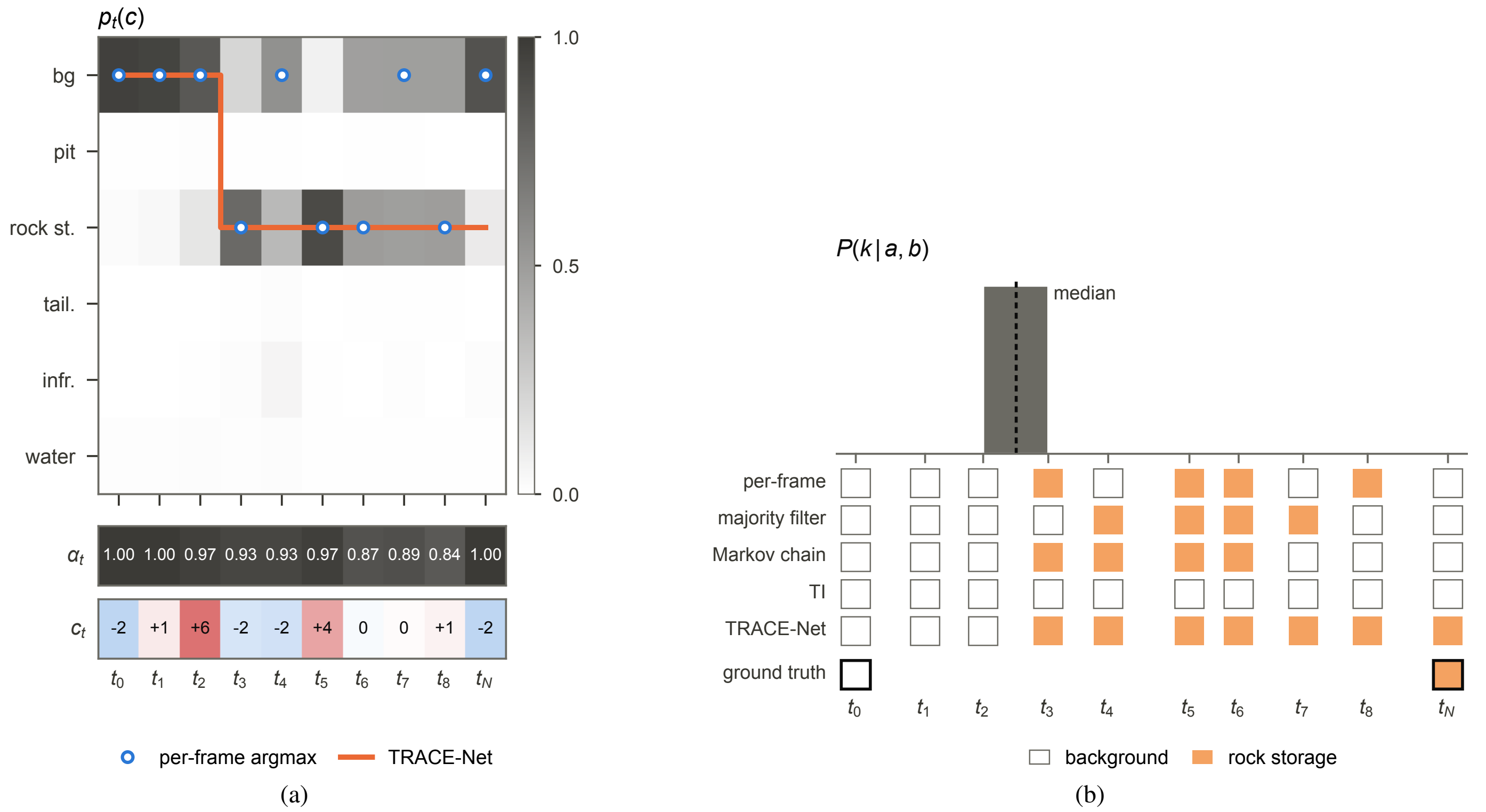}
\caption{Trajectory inference for one pixel of Kevitsa (test site). (a) Class probabilities $p_t(c)$ with the per-frame argmax (dots) and the TRACE-Net trajectory (line), agreement $\alpha_t$ and change logits $c_t$ of the anchored pairs. (b) Posterior of the change interval under TRACE-Net with its median, and the maps of the pixel from the per-frame argmax, the majority filter, the Markov chain, trajectory inference without change evidence (TI), TRACE-Net and the ground truth (endpoints only).}
\label{fig:phase2target}
\end{figure*}

Anchored inference maps each acquisition separately, so that an error at one date is not corrected by the evidence of the others, although the acquisitions are repeated observations of the same surface. Phase~2 combines the class probabilities and the change logits of all acquisitions of a pixel under the single-change assumption, on which the endpoint constraint rests: the class of a pixel changes at most once over the series (Fig.~\ref{fig:phase2target}). It requires no further training, since the Phase~1 network provides the evidence and the prior probability of change is estimated from the training annotations. Unlike a Markov chain, which scores a class sequence by its transitions between consecutive acquisitions, trajectory inference places its prior on the whole trajectory of a pixel: the classes before and after a change, taken from the endpoint transitions of the training sites, and the date of the change. It decides first what changed, over all possible dates, and then when.

\paragraph{Change prior} Let $\Pi_{\bar{t}}$ be the row-normalised endpoint transition matrix of the training annotations (computed as in Fig.~\ref{fig:bracket}a, over the training sites only), which spans their mean endpoint period of $\bar{t} = \SI{7.35}{\yr}$. The annual rate of change is $Q = (\Pi_{\bar{t}} - I)/\bar{t}$, and the transition matrix over a period of $d$ years is $\Pi_d = I + d\,Q$, with entries floored at $10^{-6}$ and rows renormalised, so that the prior follows the irregular intervals of each series.

\paragraph{Trajectory posterior} For a site without annotation, the class probabilities $p_t$, their agreement $\alpha_t$ and the change logits $c_t$ from anchored inference are the evidence of acquisition $t$. A pixel either keeps one class $a$ throughout the series or changes once, from $a$ to $b \neq a$, in the interval $k$ between acquisitions $k-1$ and $k$ ($k = 1, \ldots, N$). The posterior probability of a change is
\begin{equation}
\label{eq:switch}
\begin{split}
P(a, b, k \mid x_{0:N}) \propto{}& \pi_a\, \Pi_{H}(a,b)\, \tfrac{d_k}{H} \prod_{t<k} p_t(a)^{\omega\alpha_t} \prod_{t \geq k} p_t(b)^{\omega\alpha_t} \\
&\times \exp\!\Big(w \sum_{t} \delta_t(k)\, c_t\Big),
\end{split}
\end{equation}
and that of a constant class $a$ is proportional to $\pi_a\, \Pi_{H}(a,a) \prod_{t} p_t(a)^{\omega\alpha_t}$. Here $\pi_a$ is the class distribution of the first annotated acquisitions of the training sites, $H = t_N - t_0$ is the period of the series and $d_k$ is the length of interval $k$, so that a change is equally probable at any time. The indicator $\delta_t(k)$ is 1 when the pair that provides $c_t$ spans the change, that is, when it is the forward pair and $t<k$ or the backward pair and $t\geq k$, with the pair $(x_0, x_N)$ counted once. Because $c_t$ estimates the log-likelihood ratio of change within its pair, a change hypothesis gains $w\,c_t$ from every pair that it changes and the constant hypothesis gains nothing; each logit is limited to $\pm 8$, so that no single pair decides. The class probabilities thus determine mainly which classes a pixel has, and the change logits whether and when it changes. In the example of Fig.~\ref{fig:phase2target}, trajectory inference from the class probabilities alone maps background at every date, but the change logit is positive for the pair $(x_2, x_N)$ and negative for $(x_3, x_N)$, so that TRACE-Net infers a change from background to rock storage between $t_2$ and $t_3$, as the annotated endpoints show. The pair $(a, b)$ of highest posterior probability, summed over the intervals, is chosen first; a pixel with $a \neq b$ then takes class $b$ from the median of its posterior change date onwards. The product over acquisitions treats their evidence as conditionally independent given the trajectory, which describes errors that vary between dates but not errors that recur at every date. The evidence weight $\omega$ balances the per-frame evidence against the prior. These two inference weights, $\omega$ and the change weight $w$, are chosen on the validation sites (Section~\ref{sec:impl:phase2}).

\paragraph{Change at every time span} For two acquisitions $t_i<t_j$, we call $h=t_j-t_i$ the \emph{time span} between them. Per-frame maps detect change by comparing the classes of two dates, so that a pixel misclassified at either date appears changed. The rate of such false changes does not decrease with the time span, whereas the share of pixels that truly change grows roughly in proportion to it, so that short-term change, between consecutive acquisitions, is the most difficult to detect. Under trajectory inference, a pixel with classes $a\neq b$ and change interval $k$ is changed between $t_i$ and $t_j$ exactly when $t_i<t_k\leq t_j$. The change maps of all time spans thus follow from one trajectory per pixel and are mutually consistent, and a spurious change can occur at most once per pixel over the series, so that false changes, like true ones, accumulate with the time span.

\section{Experiments and results}
\label{sec:setup}
\label{sec:results}

\subsection{Experimental setup}
\label{sec:impl:setup}

\subsubsection{Data, protocol and implementation}
\label{sec:impl:details}
\label{sec:impl:phase2}

All networks are trained on the 13 training sites of MineNetTrace. Checkpoints and the inference weights are selected on the four validation sites, and all results are reported on the eight test sites unless stated otherwise. Trajectory inference uses an evidence weight of $\omega=4$ and a change weight of $w=0.5$, selected on the validation sites and then used unchanged for every network, for the Markov chain and on EuroMineNet. Every configuration of TRACE-Net and its variants is trained with three seeds, Phase~2 is applied to each, and results are reported as the mean. Seed standard deviations are given in \ref{app:sd}.

The intermediate acquisitions of MineNetTrace have no masks, so the timing of change is validated externally by simulating the endpoint-annotated setting on EuroMineNet \citep{yu2025euromininet}, which provides expert-verified binary mining footprints for every year from 2015 to 2024 at 133 European sites, split into 102 training, 14 validation and 17 test sites. Only the 2015 and 2024 masks are used for training and checkpoint selection, and the annual masks of the test sites serve only for scoring. The images are reduced to the nine bands of MineNetTrace, and the Phase~1 network, its training procedure and Phase~2 are those of MineNetTrace, with the change prior estimated from the EuroMineNet training annotations and the inference weights selected on MineNetTrace. As a reference for what annual masks add, the same network is also trained on every annual mask of the training sites.

The experiments were run on NVIDIA A100 and GH200 GPUs. A Phase~1 run takes about half a GPU-hour, and anchored inference over all 205 acquisitions takes about four minutes on one GPU, to which Phase~2 adds a few seconds.

\subsubsection{Evaluation metrics}
\label{sec:impl:protocol}

The metrics address the annotated endpoints, the unlabelled acquisitions and the area increments on which the production analysis relies.

\paragraph{Endpoint and change accuracy} The \emph{endpoint F1}, the per-frame macro F1 over the six classes at the annotated endpoints, computed without change masking and averaged over sites, is used as an accuracy constraint that combining the acquisitions must not lower. Change accuracy is reported with the standard SeK, the semantic change F-score $F_{\mathrm{scd}}$ and the mIoU of SECOND \citep{yang2022asymmetric}, and with a \emph{transition-space} variant of SeK, \TSeK{}. It assigns every pixel whose two classes differ to its from--to category $(a,b)$ and every other pixel to one no-change category, indexed 0, and accumulates the confusion matrix $q_{ij}$ of the reference and predicted categories. Without the no-change agreement $q_{00}$, $n=\sum_{ij}q_{ij}-q_{00}$ pixels are changed in the reference or in the map, and
\begin{equation}
\label{eq:tsek}
\begin{split}
\TSeK &= e^{\mathrm{IoU}_{\mathrm{sc}}-1}\,\frac{p_{\mathrm{o}}-p_{\mathrm{e}}}{1-p_{\mathrm{e}}},\\
p_{\mathrm{o}} &= \frac{1}{n}\sum_{i\geq 1} q_{ii},\quad
p_{\mathrm{e}} = \frac{1}{n^{2}}\sum_{i\geq 1} q'_{i+}\,q'_{+i},
\end{split}
\end{equation}
where $i$ runs over the from--to categories, $q'_{i+}$ and $q'_{+i}$ are the row and column sums over these categories, and $\mathrm{IoU}_{\mathrm{sc}}=\sum_{i,j\geq 1}q_{ij}/n$ is the intersection over union of the changed pixels; \TSeK{} is averaged over sites. Unlike SeK, \TSeK{} keeps background$\rightarrow$tailings distinct from background$\rightarrow$pit. It is one for a perfect map and zero for a map without change. Because the kappa term discounts chance agreement, it is also zero for a map with the correct change mask that assigns every change to the most frequent transition, background$\rightarrow$rock storage, for which $F_{\mathrm{scd}}$ is 0.49. The ground truth with its last mask shifted by one pixel reaches a \TSeK{} of 0.60 (Table~\ref{tab:qdad}).

\paragraph{Unlabelled acquisitions} For the endpoint-determined pixels, whose class the endpoint masks fix at every intermediate date, we report the macro F1 at the intermediate acquisitions (\emph{intermediate F1}), the \emph{pixel flip rate}, the share of these pixels whose predicted class changes at least once, and the \emph{false-change rate}, the share whose predicted class differs between two acquisitions $h$ years apart, pooled over all such pairs. Transition pixels are excluded from these three metrics. The \emph{mapped change} is the share of all annotated pixels whose predicted class differs between two acquisitions $h$ years apart; it is compared with the \emph{ground-truth change}, the share of transition pixels multiplied by $h$ and divided by the endpoint period, which assumes that the dates of change are spread uniformly. These metrics score map quality at the unlabelled acquisitions, not the timing of change. \emph{Directional consistency} $E$ divides the net change of a mapped area over the series by the sum of its absolute changes between consecutive acquisitions, so that $E=1$ when every step moves in the same direction; it is reported as the mean over the five mining classes. The \emph{change recall}, the share of transition pixels whose maps at the two endpoints differ, counts how many of the ground-truth changes a map keeps.

\paragraph{Area increments and production} Increments over the endpoint period are scored by the mean absolute increment error $|\dA|$ across the test sites, for the mining area and as the mean over the five mining classes (\emph{class mean}), which also counts area assigned to the wrong mining class. \emph{Increment fidelity} \rinc{} is the Pearson correlation across the test sites between predicted and ground-truth increments. It complements the error because production is validated by correlation: an area bias common to all sites leaves it unchanged, and if the errors are independent of the ground-truth increments and of production, the correlation of the mapped increments with production is approximately that of the ground-truth increments multiplied by \rinc{} \citep{spearman1904proof}. \emph{Retention} is the ratio of predicted to ground-truth \rprod{} for each pair of class and production indicator with ground-truth $|\rprod|>0.3$.

\paragraph{Annual masks} On EuroMineNet, the maps of 2016 to 2023 are scored by the accuracy on the changed pixels, whose 2015 and 2024 masks differ, the mining-class F1 over all ten years, the dating of change for pixels that change exactly once, as the shares dated to the correct year and within one year of it, and the \emph{annual increment fidelity}, the correlation of predicted and ground-truth annual mining-area increments within each site, averaged over sites. Annual increment fidelity scores the year-to-year variation of the mapped areas, on which the monitoring of annual production would rely. Change between any two years is scored by time span, as the F1 of the changed pixels over all pairs of years $h$ apart (\emph{change F1}), with its precision and recall, and as the false-change rate, the share of the unchanged pixels, whose masks agree at the two years, that the maps label as changed.

\subsubsection{Compared methods}
\label{sec:impl:components}

\paragraph{Published SCD architectures} The 16 published SCD architectures are trained with their own published loss functions, optimisers and schedules, and with our split, tiling, normalisation, sampler, augmentation, training budget and evaluation; unlike in the published code, background is supervised, because its area enters the production analysis. As published, twelve of them start from encoders pretrained on ImageNet (ResNet-34, EfficientNet-B0, PVTv2-B2 or VMamba), whereas CdSC, Change3D and the two HRSCD strategies are trained from random initialisation; each architecture is trained once. Each is first evaluated at the annotated endpoints, as in its publication. Because the complete method also combines the unlabelled acquisitions, trajectory inference is then applied to every baseline as well: anchored inference with the same pairs provides the class probabilities of every acquisition and, from the change output of the architecture, the change logit of each anchored pair.

\paragraph{Alternatives to trajectory inference} A Markov chain over all class sequences, with the transition matrix $\Pi_d$ between acquisitions $d$ years apart and the evidence $p_t^{\omega}$, returns, by the forward--backward algorithm \citep{rabiner1989tutorial}, the posterior class probabilities of every acquisition without the single-change assumption, so that a pixel may change any number of times. A three-frame majority filter of the per-frame maps uses no model of change. Two trivial references copy the per-frame map of an endpoint to the intermediate acquisitions: \emph{first-endpoint persistence} takes the first endpoint, so that a change falls into the last interval, and \emph{nearest endpoint} the endpoint nearer in time, so that it falls at mid-series. \emph{Break detection}, in the spirit of BFAST \citep{verbesselt2010bfast} and of temporal segmentation, fits the class-probability series of each pixel with a piecewise-constant model with zero or one break, chooses between the two by the Bayesian information criterion and assigns each segment the class of highest mean probability.

\paragraph{Variants of TRACE-Net} For Phase~1, the Siamese fusion and the change branch are removed one at a time and together; the area-consistency term, which also passes information between the dates, is removed together with the fusion. This gives four networks: \emph{no coupling}, in which no information passes between the dates; the \emph{change branch only}; the \emph{fusion only}; and TRACE-Net. The first is a Siamese segmentation network whose maps are compared after classification, and the second a multitask SCD network in the design of HRSCD strategy~4 \citep{daudt2019multitask}; both share the backbone and training procedure of TRACE-Net, so that differences between the four networks reflect the modules rather than the training. Each is scored with its per-frame maps and with its complete method, that is, trajectory inference with change evidence where the network has a change branch and without it otherwise. The bidirectional anchored inference of TRACE-Net is compared with \emph{unidirectional} anchored inference, which predicts every acquisition only in the pair with the farther endpoint, and with the plain and an entropy-weighted mean of the two predictions; unidirectional inference is also applied to every published architecture. Finally, single-factor ablations change one setting of the training procedure of TRACE-Net at a time: the backbone, the change branch, the initialisation, the normalisation, the class weighting and the tile sampling.

\paragraph{Sensitivity and number of acquisitions} The complete method is also scored with evidence weights $\omega$ from 1 to 8 and change weights $w$ from 0.25 to 2, and with the change rate of the prior multiplied by factors from 0.25 to 4. Fewer and irregularly spaced acquisitions are simulated by removing every second intermediate acquisition, a random half of them (three draws) or all but the middle one.

\paragraph{Production analysis} For the cross-mine analysis, every training and validation site is mapped by a network that was not trained on it. The 17 training and validation sites are divided into four folds of four or five sites; TRACE-Net is trained with three seeds on the non-test sites outside each fold, and its final checkpoint maps the held-out fold. The test sites keep the networks of the fixed split.

\subsection{Comparison with published SCD architectures}
\label{sec:res:zoo}

\subsubsection{Map, change and area accuracy}

Table~\ref{tab:zoo} compares the 16 published architectures with TRACE-Net at the annotated endpoints, as in their publications. In this two-date evaluation, the per-frame maps of TRACE-Net reach a higher site-averaged endpoint F1 than the best baseline, LSAFNet (0.721 against 0.681), at a similar \TSeK{}. The complete method, which adds trajectory inference, also raises the change accuracy. It has the highest endpoint F1, $F_{\mathrm{scd}}$, \TSeK{}, mIoU and overall accuracy and the highest F1 for each of the five mining classes, with the largest margins for the pit, infrastructure and water. The two HRSCD strategies, trained from random initialisation, assign no pixels to the pit, infrastructure and water.

\begin{table*}[tp]
\centering
\caption{Published SCD architectures and TRACE-Net on the test sites at the annotated endpoints. Baselines: single runs, ordered by \TSeK{}; TRACE-Net: three-seed means. Par.: parameters (millions); OA: overall accuracy; F1: site-averaged macro F1; per-class F1 pooled over pixels (infr.: infrastructure; tail.: tailings); $|\dA|$ err: mean absolute increment error (\si{\hectare}) of the mining area and the class mean (increment fidelity in Table~\ref{tab:rinc_class}). Bold: best per column except Par.\ and SeK.}
\label{tab:zoo}
\footnotesize
\setlength{\tabcolsep}{2.5pt}
\begin{tabular*}{\linewidth}{@{\extracolsep{\fill}}lrrrrrrrrrrrrrr@{}}
\toprule
 & & & & & & & & \multicolumn{5}{c}{F1 per mining class} & \multicolumn{2}{c}{$|\dA|$ err} \\
\cmidrule(lr){9-13}\cmidrule(lr){14-15}
Architecture & Par. & OA & F1 & mIoU & SeK & $F_{\mathrm{scd}}$ & \TSeK{} & pit & \colhead[r]{rock\\storage} & tail. & infr. & water & mining & \colhead[r]{class\\mean} \\
\midrule
LSAFNet & 28.4 & 0.989 & 0.681 & 0.659 & $-0.018$ & 0.384 & 0.105 & 0.727 & 0.757 & 0.843 & 0.521 & 0.393 & \textbf{27.1} & 26.9 \\
FoBa & 54.0 & 0.987 & 0.665 & 0.634 & $-0.038$ & 0.356 & 0.091 & 0.713 & 0.769 & 0.846 & 0.440 & 0.363 & 45.3 & 25.1 \\
ChangeMask & 10.6 & 0.975 & 0.672 & 0.600 & $-0.004$ & 0.295 & 0.081 & 0.702 & 0.742 & 0.847 & 0.549 & 0.454 & 64.9 & 51.4 \\
CPGNet & 55.5 & 0.977 & 0.654 & 0.609 & +0.002 & 0.313 & 0.080 & 0.736 & 0.755 & 0.857 & 0.462 & 0.383 & 48.9 & 49.2 \\
SSCD-L & 23.3 & 0.975 & 0.670 & 0.600 & $-0.001$ & 0.300 & 0.080 & 0.740 & 0.756 & 0.832 & 0.498 & 0.378 & 81.2 & 45.7 \\
BGSNet & 25.2 & 0.977 & 0.642 & 0.606 & $-0.013$ & 0.305 & 0.078 & 0.701 & 0.759 & 0.850 & 0.465 & 0.352 & 106.9 & 46.4 \\
ChangeMamba & 97.0 & 0.973 & 0.663 & 0.592 & $-0.011$ & 0.270 & 0.074 & 0.742 & 0.755 & 0.839 & 0.522 & 0.293 & 72.1 & 57.8 \\
BiSRNet & 23.4 & 0.972 & 0.641 & 0.592 & 0.000 & 0.279 & 0.073 & 0.725 & 0.747 & 0.819 & 0.478 & 0.306 & 96.8 & 61.7 \\
TED & 24.2 & 0.972 & 0.659 & 0.591 & $-0.007$ & 0.272 & 0.071 & 0.699 & 0.748 & 0.804 & 0.480 & 0.407 & 68.5 & 57.6 \\
HRSCD-str.4 & 13.7 & 0.977 & 0.371 & 0.591 & $-0.061$ & 0.252 & 0.065 & 0.000 & 0.632 & 0.767 & 0.000 & 0.000 & 106.0 & 62.6 \\
HGINet & 27.7 & 0.969 & 0.580 & 0.582 & $-0.003$ & 0.252 & 0.059 & 0.679 & 0.732 & 0.784 & 0.345 & 0.282 & 139.3 & 74.3 \\
HRSCD-str.3 & 12.8 & 0.973 & 0.359 & 0.577 & $-0.059$ & 0.216 & 0.058 & 0.000 & 0.588 & 0.721 & 0.000 & 0.000 & 196.3 & 81.0 \\
EGMSNet & 30.8 & 0.971 & 0.597 & 0.571 & $-0.029$ & 0.229 & 0.055 & 0.602 & 0.713 & 0.777 & 0.373 & 0.309 & 63.8 & 66.1 \\
GSTM-SCD & 41.9 & 0.971 & 0.519 & 0.578 & $-0.038$ & 0.211 & 0.054 & 0.485 & 0.651 & 0.801 & 0.258 & 0.248 & 76.9 & 77.2 \\
CdSC & 38.6 & 0.969 & 0.507 & 0.574 & $-0.029$ & 0.218 & 0.054 & 0.541 & 0.652 & 0.800 & 0.123 & 0.262 & 92.8 & 81.9 \\
Change3D & 9.1 & 0.964 & 0.432 & 0.532 & $-0.074$ & 0.122 & 0.035 & 0.261 & 0.439 & 0.649 & 0.102 & 0.232 & 281.2 & 84.4 \\
\midrule
\multicolumn{15}{@{}l}{TRACE-Net} \\
\quad complete method & 26.1 & \textbf{0.991} & \textbf{0.734} & \textbf{0.690} & +0.043 & \textbf{0.484} & \textbf{0.152} & \textbf{0.797} & \textbf{0.808} & \textbf{0.882} & \textbf{0.596} & \textbf{0.512} & 37.2 & \textbf{19.2} \\
\quad per-frame maps & 26.1 & 0.986 & 0.721 & 0.641 & $-0.012$ & 0.392 & 0.108 & 0.786 & 0.796 & 0.878 & 0.584 & 0.473 & 57.7 & 20.3 \\
\bottomrule
\end{tabular*}
\end{table*}

The complete method also estimates the class-mean increments most accurately, followed by FoBa and LSAFNet, whereas LSAFNet is more accurate for the mining area as a whole. Increment fidelity (Table~\ref{tab:rinc_class}) shows the same contrast between the mining area as a whole and its classes: for the mining area, several baselines exceed the fidelity of TRACE-Net, whereas averaged over the five mining classes TRACE-Net has the highest fidelity, leading for the pit, rock storage and tailings; CdSC leads for water and several baselines for infrastructure, whose increments are small at most sites.

\subsubsection{Standard and transition-space SeK}

The standard SeK is negative or close to zero for every baseline, despite overall accuracies above 0.96, and reaches +0.043 for the complete method of TRACE-Net. These values are consistent with its masking convention, which sends the from-class of a changed pixel to the no-change index whenever it is background, whereas \SI{87}{\percent} of the change in this dataset starts from background. \TSeK{}, computed from the same predictions, keeps these transitions. Across the baselines, it tracks the class-mean increment error closely (correlation $-0.94$) and ranks the methods like $F_{\mathrm{scd}}$ (rank correlation 0.95), whereas SeK does not (0.58 with $F_{\mathrm{scd}}$); unlike $F_{\mathrm{scd}}$, it gives no credit to transitions that agree by chance.

\subsubsection{Trajectory inference across architectures}
\label{sec:res:plugin}

To separate the contribution of the inference from that of the network, Table~\ref{tab:zoo_p2} applies the same trajectory inference to the published architectures, with the change evidence taken from the change output of each. Without retraining, the inference also improves their change accuracy and temporal consistency: \TSeK{}, $F_{\mathrm{scd}}$ and intermediate F1 rise and the pixel flip rate falls, \TSeK{} and $F_{\mathrm{scd}}$ typically by a third or more and the flip rate by a factor of two or more. The class-mean increment error falls for all but one architecture, and endpoint F1 changes little. The gain in temporal consistency therefore comes from the inference rather than from a particular network. With the same inference, TRACE-Net keeps the highest endpoint F1 and \TSeK{} (0.734 and 0.152, against 0.697 for ChangeMask and 0.125 for LSAFNet), so that its lead does not come from the additional acquisitions alone.

\subsection{Trajectory inference on MineNetTrace}
\label{sec:res:mapping}

\begin{table*}[tp]
\centering
\caption{Mapping without annotation on the test sites of MineNetTrace and EuroMineNet (three-seed means). MineNetTrace: endpoint F1 and \TSeK{}; $|\dA|$ err in \si{\hectare}; intermediate F1 and pixel flip rate on endpoint-determined pixels; change recall on transition pixels; directional consistency $E$ of the mapped class areas. EuroMineNet, against the annual masks: F1 over all years; acc., changed: accuracy on changed pixels at 2016 to 2023; dated: shares of single-change pixels dated to the ground-truth year and within one year of it; annual fidelity: annual increment fidelity. Dash: not applicable.}
\label{tab:mapping}
\footnotesize
\setlength{\tabcolsep}{1.5pt}
\begin{tabular*}{\linewidth}{@{\extracolsep{\fill}}lrrrrrrrrr@{\hspace{7pt}}rrrrr@{}}
\toprule
 & \multicolumn{9}{c}{MineNetTrace} & \multicolumn{5}{c}{EuroMineNet} \\
\cmidrule(lr){2-10}\cmidrule(lr){11-15}
 & & & \multicolumn{2}{c}{$|\dA|$ err} & & & & & & & & \multicolumn{2}{c}{dated} & \\
\cmidrule(lr){4-5}\cmidrule(lr){13-14}
 & F1 & \TSeK{} & mining & \colhead[r]{class\\mean} & \colhead[r]{\rinc{}\\class\\mean} & \colhead[r]{inter-\\mediate\\F1} & \colhead[r]{pixel\\flip\\rate} & \colhead[r]{change\\recall} & \colhead[r]{$E$,\\class\\mean} & \colhead[r]{F1,\\all\\years} & \colhead[r]{acc.,\\changed} & \colhead[r]{correct\\year} & \colhead[r]{within\\one\\year} & \colhead[r]{annual\\fidelity} \\
\midrule
\multicolumn{15}{@{}l}{\emph{training on the endpoint masks}} \\
\quad per-frame maps & 0.721 & 0.108 & 57.7 & 20.3 & +0.633 & 0.778 & 0.0308 & 0.548 & 0.249 & 0.873 & 0.823 & 0.392 & 0.614 & +0.02 \\
\quad first-endpoint persistence & 0.721 & 0.108 & 57.7 & 20.3 & +0.633 & 0.773 & 0.0130 & 0.548 & 1.000 & 0.845 & 0.482 & 0.035 & 0.152 & $-0.09$ \\
\quad nearest endpoint & 0.721 & 0.108 & 57.7 & 20.3 & +0.633 & 0.771 & 0.0130 & 0.548 & 1.000 & 0.861 & 0.696 & 0.087 & 0.305 & +0.10 \\
\quad three-frame majority filter & 0.721 & 0.108 & 57.7 & 20.3 & +0.633 & 0.783 & 0.0214 & 0.548 & 0.389 & 0.878 & 0.824 & 0.375 & 0.602 & +0.12 \\
\quad break detection & 0.731 & 0.114 & 56.5 & 19.7 & +0.658 & 0.787 & 0.0127 & 0.573 & 0.653 & 0.879 & 0.823 & 0.387 & 0.601 & +0.19 \\
\quad Markov-chain inference & 0.733 & 0.131 & 55.7 & 19.3 & +0.651 & 0.788 & 0.0070 & 0.491 & 0.714 & 0.881 & 0.824 & 0.373 & 0.591 & +0.18 \\
\quad trajectory inference & 0.733 & 0.131 & 54.3 & 18.8 & +0.659 & 0.789 & 0.0067 & 0.501 & 0.771 & 0.881 & 0.822 & 0.372 & 0.592 & +0.19 \\
\quad + change evidence (TRACE-Net) & 0.734 & 0.152 & 37.2 & 19.2 & +0.623 & 0.789 & 0.0055 & 0.546 & 0.862 & 0.881 & 0.827 & 0.399 & 0.618 & +0.21 \\
\multicolumn{15}{@{}l}{\emph{training on every annual mask (EuroMineNet only, reference)}} \\
\quad per-frame maps & -- & -- & -- & -- & -- & -- & -- & -- & -- & 0.867 & 0.820 & 0.385 & 0.610 & +0.13 \\
\quad trajectory inference + change evidence & -- & -- & -- & -- & -- & -- & -- & -- & -- & 0.872 & 0.826 & 0.400 & 0.631 & +0.24 \\
\bottomrule
\end{tabular*}
\end{table*}

\begin{figure*}[tp]
\centering
\input{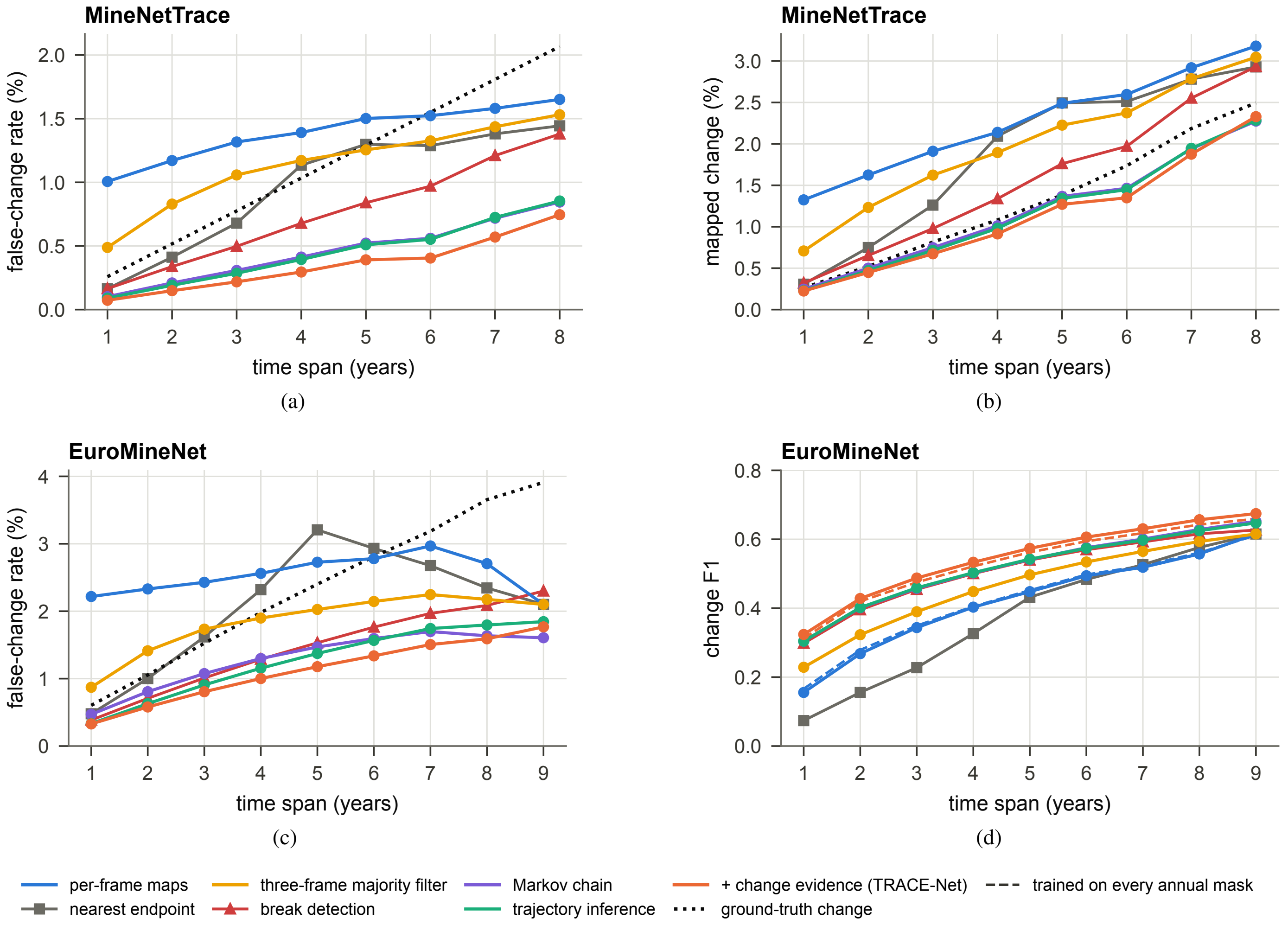}
\caption{Change detection by time span (three-seed means; Phase~1 network of TRACE-Net). (a) MineNetTrace test sites: false-change rate on endpoint-determined pixels. (b) MineNetTrace test sites: mapped change. (c) EuroMineNet test sites: false-change rate on unchanged pixels. (d) EuroMineNet test sites: change F1 (dashed: network trained on every annual mask). Dotted: ground-truth change, in (a) as the annual change rate of the test sites times the time span.}
\label{fig:horizon}
\end{figure*}

\begin{figure*}[tp]
\centering
\includegraphics[width=0.72\textwidth]{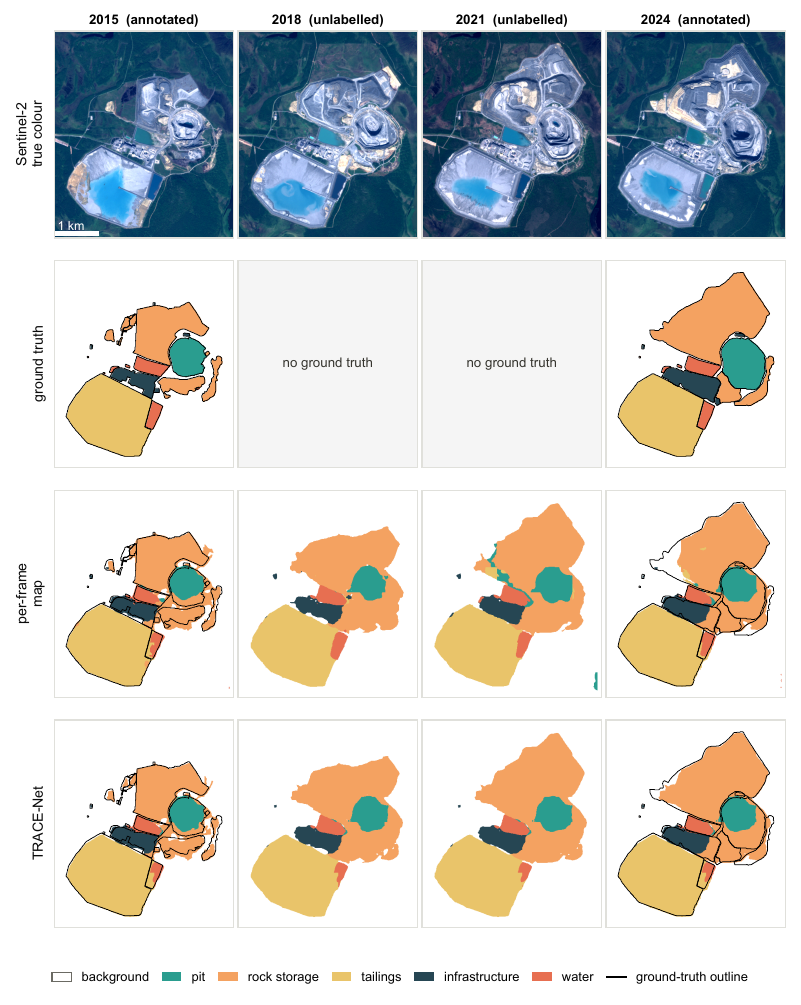}
\caption{Kevitsa (test site), cropped to the annotated mining area, at the two annotated endpoints and two unlabelled acquisitions. Rows: Sentinel-2 true-colour image; ground truth; per-frame map; TRACE-Net. Black lines: ground-truth class boundaries at the endpoints, repeated on both maps.}
\label{fig:qual}
\end{figure*}

\subsubsection{Temporal consistency and endpoint accuracy}
\label{sec:res:consistency}

Applied to the three seeds of the Phase~1 network without further training, trajectory inference makes the maps of the test sites temporally consistent (Table~\ref{tab:mapping}). The pixel flip rate on endpoint-determined pixels falls from \SI{3.1}{\percent} to \SI{0.6}{\percent}, and the class-mean directional consistency of the mapped areas rises from about 0.25 to 0.86, improving at every test site. The intermediate F1 rises at the same time, so that the maps of the unlabelled acquisitions become more accurate as well as more consistent.

This consistency is obtained without lowering the accuracy at the annotated endpoints: trajectory inference raises endpoint F1 from 0.721 to 0.734, with a gain at every test site, and \TSeK{} from 0.108 to 0.152. The gain in F1 comes from the assignment of the mining classes: with the five mining classes merged, endpoint F1 is practically the same with and without trajectory inference, so that the combined evidence resolves confusions between mining classes, such as rock storage and pit.

The change evidence mainly decides which pixels change. Without it, trajectory inference reaches the same endpoint F1 but a lower \TSeK{} and keeps fewer of the ground-truth changes between the endpoints, whereas TRACE-Net keeps as many as the per-frame maps (change recall 0.55). With the change evidence, the mining-area increment error falls from \SI{58}{\hectare} for the per-frame maps to \SI{37}{\hectare}, and the class-mean increment fidelity stays at the level of the per-frame maps.

\subsubsection{Short-term and long-term change}
\label{sec:res:horizon_mnt}

The advantage of trajectory inference is largest at short time spans (Fig.~\ref{fig:horizon}a,b). On endpoint-determined pixels, the per-frame maps change class for about \SI{1}{\percent} to \SI{1.7}{\percent} of the pixels at every time span from one to eight years (the single pair of nine years is not shown), whereas the test sites change only \SI{0.26}{\percent} of their pixels per year. Over one year, the typical interval between consecutive acquisitions, the per-frame maps therefore report about four times more false changes than the sites undergo, and their mapped change is five times the ground-truth change. TRACE-Net lowers the one-year false-change rate nearly fourteenfold, to less than a third of the annual change, and lets it grow with the time span, as true change does; the change evidence removes false changes mainly at intermediate time spans. The mapped change of TRACE-Net stays close to, and slightly below, the ground-truth change at every time span.

\subsubsection{Varying and recurring errors}

Trajectory inference corrects errors that occur at single acquisitions but not errors that recur at most acquisitions. Fig.~\ref{fig:qual} shows both kinds of error at Kevitsa. Trajectory inference corrects nearly all pixels that the per-frame maps misclassify at a single intermediate acquisition, such as the patches of rock storage and background that the per-frame map of 2021 labels as pit, and keeps the mapped pit area between 62 and \SI{66}{\hectare}, whereas it varies between 49 and \SI{93}{\hectare} across the per-frame maps. Errors that recur at every acquisition remain: both the per-frame maps and TRACE-Net assign nearly all of the expanded pit to rock storage, so that the mapped pit grows by \SI{4}{\hectare} against \SI{45}{\hectare} in the ground truth.

\subsection{External validation with annual masks}
\label{sec:res:euromine}

\subsubsection{Validity of the single-change assumption}

The annual EuroMineNet masks first test the single-change assumption (Fig.~\ref{fig:bracket}c). Over the ten years, \SI{94.9}{\percent} of the pixels labelled as mining in at least one year change at most once. Most of the \SI{5.1}{\percent} that change twice or more are mined temporarily, from background to mining and back. The assumption itself leaves little room for error: the best single-change trajectory of every pixel, chosen with knowledge of its annual masks, would be correct at \SI{99.5}{\percent} of the intermediate years on the changed pixels, so that the accuracy of trajectory inference is determined by the per-frame evidence rather than by the assumption.

\subsubsection{Accuracy at the intermediate years}

\begin{figure*}[tp]
\centering
\includegraphics[width=\textwidth]{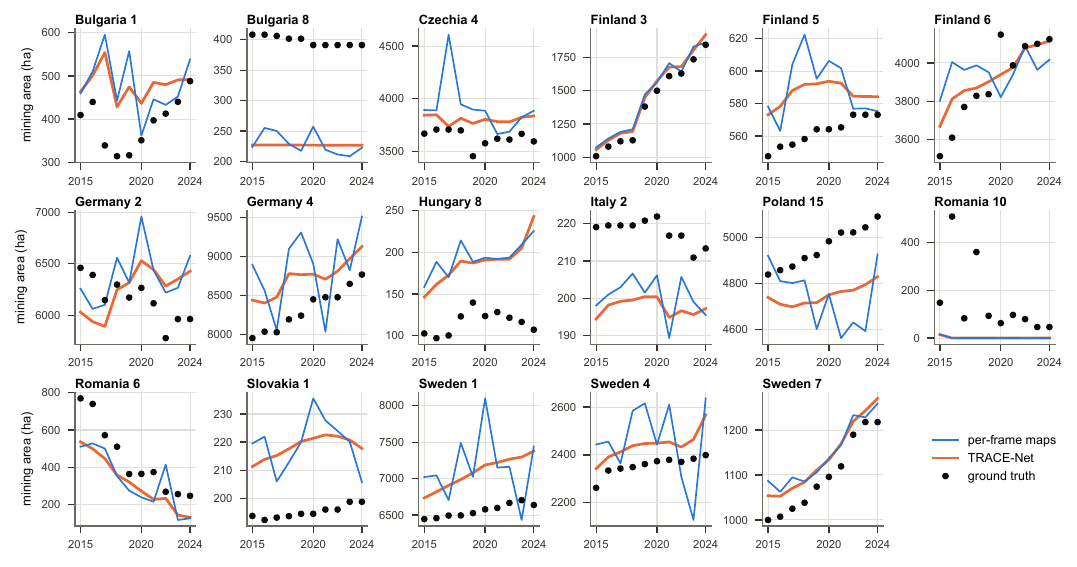}
\caption{Annual mining area at the 17 EuroMineNet test sites: ground truth, per-frame maps and TRACE-Net (three-seed means).}
\label{fig:emseries}
\end{figure*}

Table~\ref{tab:mapping} also scores the maps against the annual masks of the EuroMineNet test sites. The gains mirror those on MineNetTrace: F1 over all years rises from 0.873 to 0.881, false changes between consecutive years fall from \SI{2.2}{\percent} to \SI{0.3}{\percent} of the unchanged pixels, and the annual mining-area increments follow the ground truth more closely (Fig.~\ref{fig:emseries}), with annual increment fidelity rising from 0.02 to 0.21. On the changed pixels, TRACE-Net is slightly more accurate than the per-frame maps.

The change evidence matters most for dating reclamation, which accounts for about two fifths of the single changes at the test sites. The per-frame network recognises reclaimed land only gradually: its probability of the final class rises to 0.87 in the year of an expansion but only to 0.44 in the year of a reclamation. From these probabilities alone, trajectory inference dates fewer changes to the correct year than the per-frame maps, but the change evidence restores the correct dates. TRACE-Net thus dates changes about as accurately as the per-frame maps, \SI{40}{\percent} to the ground-truth year and \SI{62}{\percent} within one year of it, while removing most of their false changes.

\subsubsection{Change detection across time spans}
\label{sec:res:horizon_em}

The annual masks score the change between every pair of years (Fig.~\ref{fig:horizon}c,d). Short-term change is the most difficult to detect. Over one year, only \SI{0.6}{\percent} of the test-site pixels truly change, but the per-frame maps flag \SI{2.2}{\percent} of the unchanged pixels as changed, so that only about one in ten detected changes is real and the change F1 is only 0.16, against 0.62 over the full period. TRACE-Net keeps the false changes below the true changes at every time span and doubles the one-year change F1, to 0.32. At one year it gains mainly in precision, which rises from 0.10 to 0.36, at a recall of 0.30 against 0.39 for the per-frame maps; over the full period it detects more of the true changes.

\subsubsection{Annual masks as training data}
\label{sec:res:annual}

Training on the 2015 and 2024 masks alone performs about as well as training on all ten annual masks of the same sites (Table~\ref{tab:mapping}): F1 over all years is marginally higher, for the per-frame maps and for TRACE-Net, and the one-year change F1 is nearly the same (Fig.~\ref{fig:horizon}d). After trajectory inference, the additional annual masks improve the dating of changes and annual increment fidelity only slightly.

\subsection{Alternatives to trajectory inference}
\label{sec:res:alternatives}

Table~\ref{tab:mapping} and Fig.~\ref{fig:horizon} also compare trajectory inference with other ways of combining the acquisitions. Copying an endpoint map to the intermediate acquisitions places every change in one interval, so that the two trivial references reach a directional consistency of one. They nevertheless leave about twice the pixel flip rate of trajectory inference, lower the intermediate F1 below that of the per-frame maps and, on EuroMineNet, date at most a third of the changes within one year. The three-frame majority filter, which uses no model of change, lowers the pixel flip rate by about a third and halves the false changes at one year, but removes little at longer time spans and leaves the endpoint maps unchanged.

Break detection allows at most one change but has neither a prior on change nor change evidence. It keeps slightly more of the ground-truth changes than TRACE-Net, but at more than twice its flip rate and with a lower \TSeK{}, and at eight years it keeps most of the false changes of the per-frame maps, because every break in the class probabilities becomes a change. The Markov chain has a prior on change but allows any number of changes; it reaches the same endpoint F1 and \TSeK{} as trajectory inference without change evidence but a lower directional consistency, because it still lets a mapped area move back and forth. On EuroMineNet, both reach a one-year change F1 of about 0.30, as does trajectory inference without change evidence, against 0.23 for the majority filter. Of these methods, only trajectory inference uses the change evidence, which raises \TSeK{} and the one-year change F1 further. The single-change assumption and the change evidence therefore add more than temporal smoothing alone.

\subsection{Validation against reported production}
\label{sec:res:prod}

\begin{figure*}[tp]
\centering
\includegraphics[width=\textwidth]{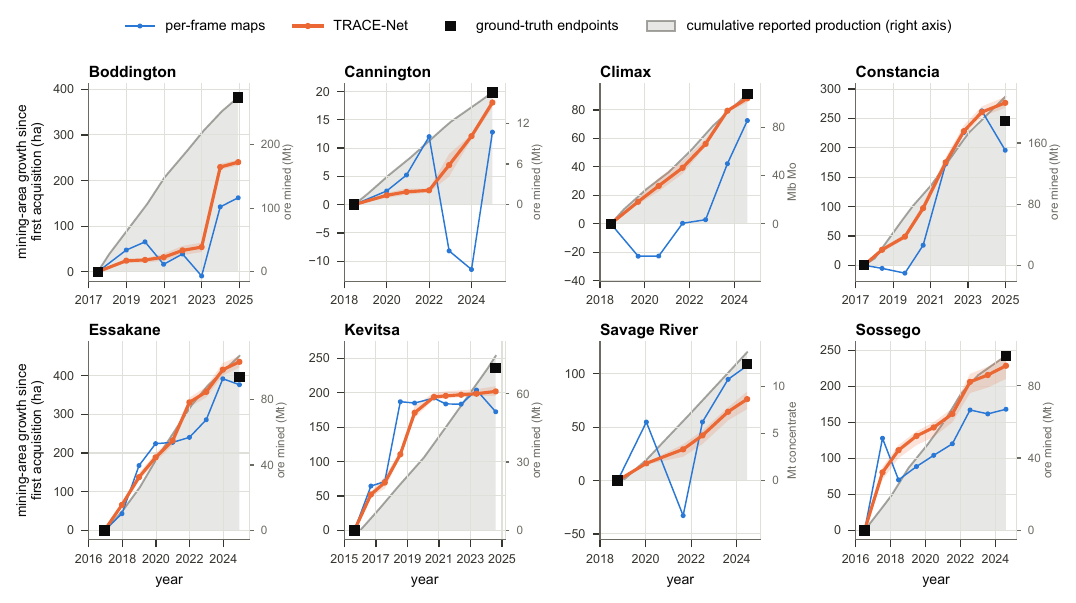}
\caption{Growth of the mapped mining area since the first acquisition at the eight test sites: per-frame maps (thin line), TRACE-Net (thick line; seed mean, range shaded) and ground-truth endpoints (squares), with cumulative reported production since the first acquisition (grey shading, right axis; best-covered indicator).}
\label{fig:traj}
\end{figure*}

Fig.~\ref{fig:traj} compares the mapped growth of the mining area at the test sites with their cumulative reported production. The mining area mapped by TRACE-Net grows steadily, mostly in step with cumulative production, and ends within about \SI{40}{\hectare} of the ground-truth growth at every test site except Boddington, whereas the per-frame maps fluctuate by tens of hectares between acquisitions. The pit is mapped less reliably: at Boddington, Kevitsa and Savage River the mapped pit grows much less than in the ground truth, because most of the new pit, cut partly into former waste-rock dumps, is mapped as rock storage at every acquisition.

Table~\ref{tab:retention} repeats the cross-mine analysis of the annotated increments with predicted increments, with every site mapped by a network that was not trained on it. For the 18 pairs of class and production indicator with ground-truth $|\rprod|>0.3$, TRACE-Net retains a median of 0.99 of the ground-truth correlation, and between about 0.6 and 1.3 for every pair except water against main-commodity output and the two infrastructure pairs. The per-frame maps retain less and reverse all but one of the five main-commodity correlations. The mapped mining area tracks ore mined as closely as the ground truth ($\rprod=+0.71$ against $+0.68$), and the size-adjusted association of the mapped pit area remains positive but is weaker than in the ground truth ($\rpart=+0.52$ against $+0.88$). The mapped area increments thus retain most of the ground-truth relationships with production across mines, and accumulated increments follow the multi-year expansion rate at most mines.

Year-to-year fluctuations in production, by contrast, are too small to be resolved by the maps. Several of the mines operate near capacity: within a mine, a fluctuation of ore mined by one standard deviation corresponds to only about \SI{3}{\hectare} of pit growth per year, judging by the size-adjusted relationship across mines, whereas shifting the outline of the pit by one pixel changes its annual increment by about \SI{10}{\hectare}. Mean expansion rates accumulate over time, so that the multi-year growth of the mapped areas, but not yet their annual fluctuations, can be related to production.

\subsection{Analysis of the components}
\label{sec:res:components}

\subsubsection{Fusion and change branch}
\label{sec:res:design}

\begin{table*}[tp]
\centering
\caption{Fusion and change branch on the test sites (three-seed means). TI: trajectory inference without change evidence; $|\dA|$ err: class-mean increment error (\si{\hectare}).}
\label{tab:design}
\footnotesize
\setlength{\tabcolsep}{3.5pt}
\begin{tabular*}{\linewidth}{@{\extracolsep{\fill}}lrrrrrrrrr@{}}
\toprule
 & \multicolumn{2}{c}{per-frame maps} & TI & \multicolumn{6}{c}{complete method} \\
\cmidrule(lr){2-3}\cmidrule(lr){4-4}\cmidrule(lr){5-10}
Variant & F1 & \TSeK{} & \TSeK{} & F1 & \TSeK{} & $F_{\mathrm{scd}}$ & \colhead[r]{\rinc{}\\mining} & \colhead[r]{\rinc{}\\class mean} & \colhead[r]{$|\dA|$ err\\class mean} \\
\midrule
no coupling & 0.709 & 0.092 & 0.104 & 0.732 & 0.104 & 0.369 & +0.949 & +0.403 & 42.0 \\
change branch only & 0.703 & 0.087 & 0.109 & 0.730 & 0.131 & 0.439 & +0.960 & +0.415 & 33.8 \\
fusion only & 0.722 & 0.106 & 0.125 & 0.734 & 0.125 & 0.431 & +0.876 & +0.657 & 20.1 \\
fusion + change branch (TRACE-Net) & 0.721 & 0.108 & 0.131 & 0.734 & 0.152 & 0.484 & +0.912 & +0.623 & 19.2 \\
\bottomrule
\end{tabular*}
\end{table*}

The two components contribute complementary gains (Table~\ref{tab:design}): each raises \TSeK{} on its own, together they raise it from 0.104 to 0.152, and each acts on a different part of the evidence. The fusion improves the class evidence of each date: it halves the class-mean increment error and raises the class-mean increment fidelity from 0.40 to 0.66, while the networks without fusion have a higher fidelity for the mining area as a whole. The change branch improves the change evidence: its own per-frame maps are no better, and its contribution appears when its change logits enter trajectory inference, which raises \TSeK{} by about 0.02. Endpoint F1 varies little across the four variants.

On EuroMineNet, the two components act on the precision and the recall of change detection. Compared with the network without fusion or change branch, TRACE-Net raises the precision at every time span at nearly the same recall, and so achieves a higher change F1 at every time span (0.32 against 0.26 at one year). The anchored pairs of the test sites show how: Siamese fusion nearly halves the false changes of the per-frame maps, but an error at one date then recurs at the other more often, and more true changes are missed; in TRACE-Net, the change evidence recovers part of these missed changes.

\subsubsection{Inference and training settings}
\label{sec:res:settings}

Appendix~A reports the remaining analyses of TRACE-Net: bidirectional anchored inference across the published architectures (Table~\ref{tab:zoo_p2}), the inference variants and settings of TRACE-Net (Table~\ref{tab:infer}) and the single-factor ablations of its training procedure (Table~\ref{tab:ablation_main}).

Bidirectional anchored inference gives every intermediate acquisition two predictions in different temporal contexts, and their agreement tells trajectory inference how far to trust the acquisition. Its benefit therefore grows with how often the two predictions differ: across the published architectures, it lowers the pixel flip rate of trajectory inference more where they differ more often (correlation 0.95) and raises \TSeK{} by up to a fifth. For TRACE-Net, whose Siamese fusion already gives each date the context of the other, the two predictions differ at only about \SI{1}{\percent} of the pixels and the gains are smaller; they are clearest on EuroMineNet, where annual increment fidelity rises from 0.17 to 0.21. The agreement weighting reaches the accuracy of the plain mean of the two predictions with a slightly lower pixel flip rate.

The complete method depends little on its inference settings. On MineNetTrace, \TSeK{} stays between 0.141 and 0.152 over the tested evidence and change weights, above that of trajectory inference without change evidence, and scaling the change rate of the prior leaves it practically unchanged; on EuroMineNet, F1 over all years varies by less than 0.003. Fewer and irregularly spaced acquisitions reduce the gains only gradually: with every second or a random half of the intermediate acquisitions removed, TRACE-Net keeps nearly its full \TSeK{}, and with a single intermediate acquisition it still raises \TSeK{} from 0.108 to 0.143.

Of the training settings, the backbone and the initialisation matter most. U-Net with ResNet-34 \citep{ronneberger2015unet} and DeepLabv3+ with ResNet-50 \citep{chen2018encoder} lower both \TSeK{} and endpoint F1, whereas UPerNet with ConvNeXt-T \citep{xiao2018unified,liu2022convnet} reaches nearly the accuracy of TRACE-Net with SegFormer-B2, so that the design also works with a ConvNeXt backbone. A randomly initialised encoder lowers \TSeK{} to 0.059 and endpoint F1 to 0.415, so that ImageNet pretraining transfers to the nine-band input. The lightweight change branch provides the change evidence at least as well as an SCD-style change decoder that compares the encoder features of both dates at every scale. The normalisation, the class weighting and the tile sampling change endpoint F1 or \TSeK{} by smaller amounts, and the settings of TRACE-Net give the highest values of both among these variants.

\section{Discussion}
\label{sec:discussion}

\subsection{Trajectory inference and short-term change}
\label{sec:disc:objectives}

Our results show how TRACE-Net improves the mapping of change between consecutive years. The change of a mine within one year is smaller than the error of a single map: at Boddington, for example, the mining area of about \SI{3500}{\hectare} grows by about \SI{50}{\hectare} per year, so that a \SI{1}{\percent} error in its mapped area amounts to about two thirds of the annual change. Two kinds of error then behave in opposite ways. An error that varies between dates produces a false change whenever two maps are compared, but it is removed when the dates are combined. An error that is the same at every date produces no false change, because it appears in both maps, but combining the dates cannot remove it either. Trajectory inference removes errors of the first kind, such as boundary errors at single misregistered acquisitions, so that the remaining false changes grow with the time between the compared dates, as true change does. Its change evidence also recovers changes that the class probabilities alone leave uncertain. On EuroMineNet, this matters most for reclamation, because the network recognises reclaimed land only gradually. Errors of the second kind remain and now limit the accuracy: at Boddington, Kevitsa and Savage River, most of the new pit is mapped as rock storage at every acquisition. Since on EuroMineNet the single-change trajectory itself would be correct at nearly every intermediate year, further gains are expected mainly from more accurate per-frame evidence rather than from more elaborate temporal models.

\subsection{Unlabelled acquisitions and annotation strategy}
\label{sec:disc:unlabelled}

The experiments also show how the unlabelled acquisitions add value. As repeated observations of each pixel, combined under the single-change assumption, they improve the maps without further training and at negligible cost, also for other SCD architectures. The single-change assumption and the change evidence are central to the benefit: a Markov chain that allows any number of changes leaves the mapped areas moving back and forth, and break detection, without change evidence or a prior on change, turns every break in the class probabilities into a change. Additional annotated dates, by comparison, add little as training data: under a split by site, the two endpoint masks already provide nearly all the training information that the annual masks contain for binary mining footprints. For new datasets, this suggests spending annotation effort on the endpoints of more sites rather than on the intermediate dates of the same sites, and reserving annual masks for evaluation, where they measure the timing of change that endpoint masks cannot.

\subsection{Temporal consistency from inference rather than from coupling}
\label{sec:disc:consistency}

The comparison of the four Phase~1 variants, with and without Siamese fusion and change branch (Table~\ref{tab:design}), indicates where temporal consistency is best obtained. Coupling the dates inside the network, so that each date draws on the features of the other, makes their maps alike whether or not they should be: Siamese fusion removes false changes, but it also makes an error at one date recur at the other and hides true changes. Trajectory inference instead obtains consistency from the whole series and from a model of how the classes change, and removes false changes while keeping the changes between the endpoints. Coupling remains useful for the evidence of each date: fusion raises the precision of change detection and the class-mean increment fidelity, and the change evidence of the change branch recovers part of the changes that fusion misses. TRACE-Net therefore uses coupling to improve the evidence of each date and trajectory inference to make the dates consistent.

\subsection{Evaluation at unlabelled dates}
\label{sec:disc:measurement}

Two findings concern how maps of a time series should be evaluated. First, the standard metrics hide the temporal behaviour of a method. Where change is dominated by conversion from a background class, as also in urban expansion and deforestation, the masking convention of SeK removes most of the transition information, and we suggest that such datasets report SeK and \TSeK{} side by side. Endpoint accuracy says equally little about time: the four Phase~1 variants reach nearly the same endpoint F1 with their complete method but differ strongly in change accuracy and increment fidelity. Second, metrics derived from the endpoints, such as the pixel flip rate, and metrics that need no labels, such as directional consistency, show whether a map is stable where the annotations imply that it should be, and whether it keeps the changes between the endpoints, but they cannot score the timing of change; simulating the endpoint-annotated setting on a dataset with annual masks measures the timing directly.

\subsection{Mapped areas and reported production}
\label{sec:disc:production}

The mapped area increments are broadly consistent with reported production across mines, and the association of pit expansion with ore mined, adjusted for initial mine size, remains positive. This extends the whole-mine, single-epoch relationship of \citet{werner2020global} to individual mine components and to mapped increments. Within a mine, however, the change in area that follows from year-to-year fluctuations in production is smaller than the boundary error of a \SI{10}{\metre} sensor. Surface change and production can also be offset in time: waste rock is stripped to expose ore before the ore is mined, so that a pit may widen before its ore is reported, and tailings facilities and waste-rock dumps expand in construction stages rather than with each year's output. Multi-year expansion rates, whose signal accumulates over time and which are less affected by such lags, can therefore be compared with production, whereas monitoring annual production from the maps will require denser within-year sampling, finer spatial resolution, elevation data that capture volume as well as area, or mines whose production changes more strongly.

\subsection{Limitations}
\label{sec:disc:limitations}

Several limitations should guide interpretation. The reported hectare values are mapped areas, obtained by counting the pixels of each class. Because omission and commission errors need not cancel, such counts can be biased; unbiased area estimates would require reference labels at a probability sample of pixels and an estimator that corrects the mapped areas for the errors observed in this sample \citep{olofsson2014good,stehman2019key}. Label noise at class boundaries and the residual misregistration of about one pixel at one pair of consecutive acquisitions in ten add to the boundary error of the area increments. All acquisitions were also selected cloud-free at approximately annual spacing, so that the method remains to be tested on cloud-affected and denser series.

Trajectory inference assumes that a pixel changes at most once. The assumption holds for \SI{94.9}{\percent} of the EuroMineNet pixels that are ever mining but fails more often at sites with repeated disturbance and recovery. Where a pixel is mined and then recovers within the series, the recovery could be mapped separately, by a network trained for it or by vegetation-based time-series analysis \citep{nwazelibe2026}, so that each step involves a single change. The inference also cannot remove errors that recur at every acquisition. On MineNetTrace the timing of change can be scored only indirectly; its direct validation relies on EuroMineNet, which is European and binary, and annotating intermediate acquisitions at several test sites would allow the timing of six-class transitions to be checked directly.

The comparisons rest on three seeds per configuration, single runs of the published architectures with their own training settings and four validation sites. The cross-mine production analysis rests on at most twelve mines per indicator, and its main indicator and responses were chosen after inspection of the annotated correlations, so that it is exploratory.

\section{Conclusions}
\label{sec:conclusion}

Consistent annual maps of mine growth are needed to track the global footprint of society's resource consumption, but the change of a mine between consecutive years is commonly smaller than the error of a single map, and reference masks usually exist for only one or two dates per site. This study showed that both challenges can be addressed by treating the images between the first and last dates as repeated observations of one change trajectory, constrained by the known or predicted classes at these dates. For this purpose, we introduced MineNetTrace, a global Sentinel-2 benchmark with endpoint masks and production records, and TRACE-Net, a novel trajectory inference approach.

The experiments lead to three conclusions. First, most errors that limit annual mapping vary between dates, so combining all acquisitions of a pixel helps remove them. Without further training, trajectory inference reduces one-year false changes nearly fourteenfold on MineNetTrace while keeping as many ground-truth changes as the per-frame maps, and TRACE-Net doubles the one-year change F1 on EuroMineNet. The same inference also improves the maps of other published semantic change detection architectures.

Second, the information needed to train the mapping is concentrated at the endpoints: on EuroMineNet, two masks per site train a network about as accurate as one trained on annual masks. For new datasets, annotation effort is therefore better spent on the endpoints of more sites than on the intermediate dates of the same sites, with annual masks used to evaluate the timing of change.

Third, maps of unlabelled acquisitions can be evaluated without intermediate references. Metrics derived from the endpoints measure the temporal consistency of the intermediate maps, and a transition-space Separated Kappa keeps the changes from background that the standard metric discards. Reported production provides an independent physical reference: across mines, the expansion of the mining area tracks ore mined, and that of open pits does so also after controlling for mine size; the mapped area increments are broadly consistent with reported production.

These results support the automated annual semantic segmentation of mines from the Sentinel-2 archive, with training sites annotated at only two dates. Such maps can track the expansion of open pits, rock storage and tailings facilities globally. They can thus help to quantify the environmental footprint of mining, to assess whether planning and regulatory requirements are met, and to manage each mine component according to its risks.

\section*{Data and code availability} The MineNetTrace imagery, the endpoint annotations, the fixed site split, the harmonised production table with its per-value quality flags, the training and analysis code, the baseline suite and the per-run evaluation artefacts of all runs reported here are available at
\url{https://github.com/EricYu97/MineNetTrace}.
The annotations and the production table are released under the Creative Commons Attribution 4.0 International licence and the code under the MIT licence; the imagery contains modified Copernicus Sentinel data (2015--2024) and is redistributed under the terms of use of Copernicus Sentinel data. Operator production figures were compiled from publicly available company reports.

\section*{Declaration of competing interest} The authors declare that they have no known competing financial interests or personal relationships that could have appeared to influence the work reported in this paper.

\section*{Acknowledgements}
Weikang Yu's work was supported by the 3D-ABC project as part of the Helmholtz Foundation Model Initiative. The authors would also like to thank the European Regional Development Fund and the Land of Saxony for providing the high-specification NVIDIA A100 GPU servers that we used in our experiments.

\FloatBarrier
\bibliographystyle{elsarticle-harv}
\bibliography{MineNetSCD_refs}
\balance      %

\onecolumn
\appendix
\setcounter{table}{0}
\setcounter{figure}{0}
\renewcommand{\theHtable}{\Alph{section}.\arabic{table}}
\renewcommand{\theHfigure}{\Alph{section}.\arabic{figure}}
\section{Supplementary results}
\label{app:results}
\setcounter{table}{0}

\begin{figure}[!htbp]
\centering
\includegraphics[width=\textwidth]{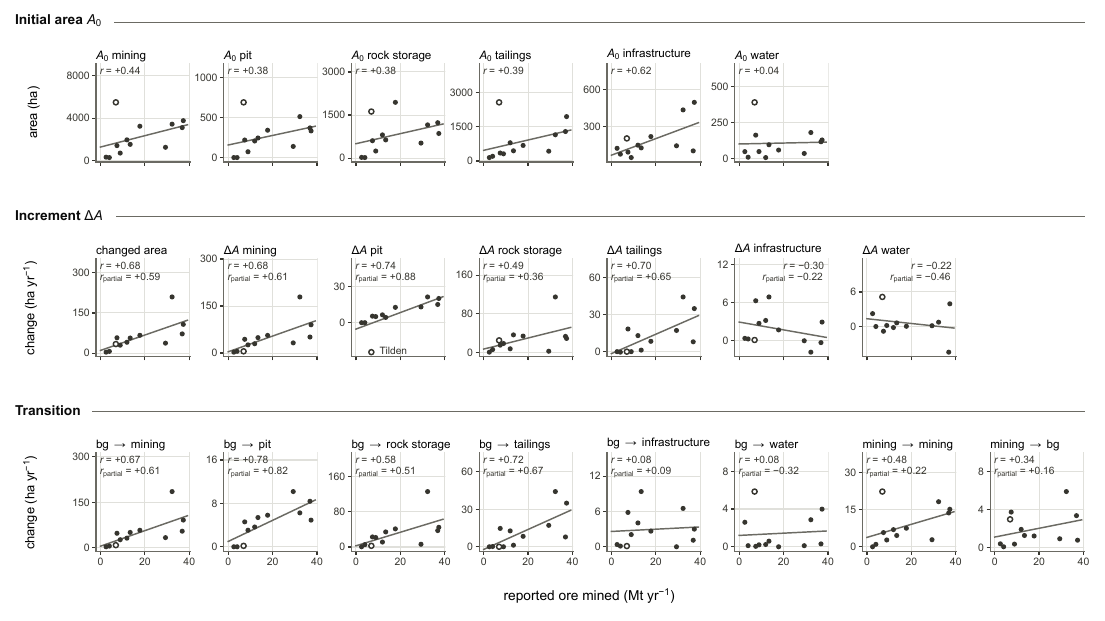}
\caption{Map-derived variables of Fig.~\ref{fig:prelim} against reported ore mined, one point per mine ($n=12$), from the annotated masks, with the least-squares line (open marker: Tilden). bg: background.}
\label{fig:prelim_scatter}
\end{figure}

This appendix gives the complete results behind statements of the main text: the cross-mine correlations of the annotated increments with their intervals (Table~\ref{tab:prodci}) and each map-derived variable against ore mined for every mine (Fig.~\ref{fig:prelim_scatter}), the change metrics of reference maps (Table~\ref{tab:qdad}), increment fidelity for every class (Table~\ref{tab:rinc_class}), trajectory inference and bidirectional anchored inference applied to the published architectures (Table~\ref{tab:zoo_p2}), the retention of the cross-mine relationships (Table~\ref{tab:retention}), and the inference variants and settings of TRACE-Net (Table~\ref{tab:infer}) and its single-factor ablations (Table~\ref{tab:ablation_main}).

\begin{table*}[!htbp]
\centering
\caption{Cross-mine correlation of the annotated area increments with production: Pearson \rprod{} and partial \rpart{} given the initial mining area, with 95\,\% intervals (Fisher $z$) and Holm-adjusted $p$-values over the 144 correlations of Fig.~\ref{fig:prelim}, and the Spearman correlation $\rho$ with its size-adjusted counterpart $\rho_{\mathrm{partial}}$ (partial correlation of the ranks).}
\label{tab:prodci}
\footnotesize
\setlength{\tabcolsep}{3pt}
\begin{tabular*}{\linewidth}{@{\extracolsep{\fill}}lrrrrrrrrr@{}}
\toprule
 & $n$ & \rprod{} & interval & Holm $p$ & \rpart{} & interval & Holm $p$ & $\rho$ & $\rho_{\mathrm{partial}}$ \\
\midrule
\multicolumn{10}{@{}l}{\emph{ore mined}} \\
\quad mining & 12 & +0.68 & +0.17 to +0.90 & 1.00 & +0.61 & +0.02 to +0.89 & 1.00 & +0.88 & +0.82 \\
\quad pit & 12 & +0.74 & +0.28 to +0.92 & 0.85 & +0.88 & +0.60 to +0.97 & 0.045 & +0.91 & +0.89 \\
\quad rock storage & 12 & +0.49 & $-0.12$ to +0.83 & 1.00 & +0.36 & $-0.31$ to +0.79 & 1.00 & +0.59 & +0.35 \\
\quad tailings & 12 & +0.70 & +0.21 to +0.91 & 1.00 & +0.65 & +0.08 to +0.90 & 1.00 & +0.72 & +0.63 \\
\quad infrastructure & 12 & $-0.30$ & $-0.75$ to +0.33 & 1.00 & $-0.22$ & $-0.72$ to +0.44 & 1.00 & $-0.20$ & $-0.14$ \\
\quad water & 12 & $-0.22$ & $-0.71$ to +0.40 & 1.00 & $-0.46$ & $-0.83$ to +0.19 & 1.00 & $-0.06$ & $-0.33$ \\
\multicolumn{10}{@{}l}{\emph{pit, other tonnage indicators}} \\
\quad ore processed & 11 & +0.74 & +0.26 to +0.93 & 1.00 & +0.24 & $-0.46$ to +0.76 & 1.00 & +0.77 & +0.39 \\
\quad total material mined & 6 & +0.78 & $-0.08$ to +0.97 & 1.00 & $-0.26$ & $-0.93$ to +0.81 & 1.00 & +0.60 & $-0.28$ \\
\bottomrule
\end{tabular*}
\end{table*}

\begin{table}[!htbp]
\centering
\caption{Change metrics of reference maps derived from the annotations of the test sites (site averages).}
\label{tab:qdad}
\footnotesize
\setlength{\tabcolsep}{4pt}
\begin{tabular}{lrrr}
\toprule
 & \TSeK{} & $F_{\mathrm{scd}}$ & SeK \\
\midrule
ground truth (perfect map) & 1.000 & 1.000 & +1.000 \\
no change & 0.000 & 0.000 & 0.000 \\
correct change mask, most frequent transition & 0.000 & 0.487 & $-0.028$ \\
ground truth, last mask shifted by one pixel & 0.595 & 0.922 & +0.685 \\
\bottomrule
\end{tabular}
\end{table}

\begin{table*}[!htbp]
\centering
\caption{Increment fidelity \rinc{} on the test sites for every mining class, their mean and the mining area. Component analysis and TRACE-Net: three-seed means; published architectures: single runs, in the order of Table~\ref{tab:zoo}. n/a: the predicted increments of the class do not vary across sites.}
\label{tab:rinc_class}
\footnotesize
\setlength{\tabcolsep}{4pt}
\begin{tabular*}{\linewidth}{@{\extracolsep{\fill}}llrrrrrrr@{}}
\toprule
 & & pit & rock storage & tailings & infrastructure & water & class mean & mining \\
\midrule
\multicolumn{9}{@{}l}{\emph{Components of TRACE-Net (per-frame maps), three-seed means}} \\
\multicolumn{2}{@{}l}{\quad no coupling} & $-0.251$ & +0.835 & +0.580 & +0.098 & +0.704 & +0.393 & +0.901 \\
\multicolumn{2}{@{}l}{\quad change branch only} & $-0.151$ & +0.891 & +0.433 & +0.025 & +0.445 & +0.329 & +0.853 \\
\multicolumn{2}{@{}l}{\quad fusion only} & +0.280 & +0.932 & +0.781 & +0.161 & +0.827 & +0.596 & +0.821 \\
\multicolumn{2}{@{}l}{\quad fusion + change branch} & +0.273 & +0.969 & +0.858 & +0.126 & +0.941 & +0.633 & +0.848 \\
\addlinespace
\multicolumn{2}{@{}l}{TRACE-Net, complete method} & +0.504 & +0.952 & +0.918 & $-0.135$ & +0.875 & +0.623 & +0.912 \\
\multicolumn{2}{@{}l}{TRACE-Net, trajectory inference without change evidence} & +0.433 & +0.959 & +0.918 & +0.081 & +0.907 & +0.659 & +0.885 \\
\midrule
\multicolumn{9}{@{}l}{\emph{Published architectures}} \\
\multicolumn{2}{@{}l}{LSAFNet} & +0.271 & +0.888 & +0.851 & $-0.391$ & +0.330 & +0.390 & +0.976 \\
\multicolumn{2}{@{}l}{FoBa} & $-0.276$ & +0.889 & +0.866 & $-0.130$ & +0.774 & +0.425 & +0.954 \\
\multicolumn{2}{@{}l}{ChangeMask} & $-0.435$ & +0.896 & +0.612 & +0.525 & +0.719 & +0.464 & +0.849 \\
\multicolumn{2}{@{}l}{CPGNet} & $-0.306$ & +0.819 & +0.828 & +0.782 & +0.432 & +0.511 & +0.976 \\
\multicolumn{2}{@{}l}{SSCD-L} & $-0.500$ & +0.920 & +0.632 & +0.519 & +0.555 & +0.425 & +0.918 \\
\multicolumn{2}{@{}l}{BGSNet} & $-0.185$ & +0.860 & +0.533 & +0.564 & +0.484 & +0.451 & +0.849 \\
\multicolumn{2}{@{}l}{ChangeMamba} & $-0.134$ & +0.863 & +0.264 & +0.607 & +0.478 & +0.415 & +0.824 \\
\multicolumn{2}{@{}l}{BiSRNet} & $-0.365$ & +0.891 & +0.371 & +0.542 & +0.668 & +0.421 & +0.877 \\
\multicolumn{2}{@{}l}{TED} & $-0.495$ & +0.920 & +0.206 & $-0.193$ & +0.674 & +0.222 & +0.879 \\
\multicolumn{2}{@{}l}{HRSCD-str.4} & n/a & +0.929 & +0.432 & $-0.305$ & n/a & n/a & +0.934 \\
\multicolumn{2}{@{}l}{HGINet} & $-0.299$ & +0.780 & +0.619 & +0.022 & +0.558 & +0.336 & +0.841 \\
\multicolumn{2}{@{}l}{HRSCD-str.3} & n/a & +0.259 & +0.090 & n/a & n/a & n/a & +0.267 \\
\multicolumn{2}{@{}l}{EGMSNet} & $-0.014$ & +0.839 & +0.320 & $-0.391$ & +0.726 & +0.296 & +0.948 \\
\multicolumn{2}{@{}l}{GSTM-SCD} & $-0.256$ & +0.736 & +0.315 & +0.275 & +0.403 & +0.294 & +0.929 \\
\multicolumn{2}{@{}l}{CdSC} & $-0.679$ & +0.562 & +0.524 & +0.557 & +0.977 & +0.388 & +0.656 \\
\multicolumn{2}{@{}l}{Change3D} & $-0.407$ & +0.408 & +0.519 & $-0.278$ & +0.024 & +0.053 & +0.083 \\
\bottomrule
\end{tabular*}
\end{table*}

\begin{table*}[!htbp]
\centering
\caption{Trajectory inference applied to the published architectures on the test sites (order of Table~\ref{tab:zoo}). PF: per-frame maps from anchored inference; TN: trajectory inference with the change output of the network and bidirectional anchored inference; uni.: the same with unidirectional anchored inference; differ: share of the endpoint-determined pixels at the intermediate acquisitions whose forward and backward predictions differ in class; $|\dA|$ err: class-mean increment error (\si{\hectare}).}
\label{tab:zoo_p2}
\footnotesize
\setlength{\tabcolsep}{2pt}
\begin{tabular*}{\linewidth}{@{\extracolsep{\fill}}lrrrrrrrrrrrrrrr@{}}
\toprule
 & \multicolumn{2}{c}{F1} & \multicolumn{3}{c}{\TSeK{}} & \multicolumn{2}{c}{$F_{\mathrm{scd}}$} & \multicolumn{2}{c}{$|\dA|$ err} & \multicolumn{2}{c}{intermediate F1} & \multicolumn{3}{c}{pixel flip rate} & differ \\
\cmidrule(lr){2-3}\cmidrule(lr){4-6}\cmidrule(lr){7-8}\cmidrule(lr){9-10}\cmidrule(lr){11-12}\cmidrule(lr){13-15}
Architecture & PF & TN & PF & uni. & TN & PF & TN & PF & TN & PF & TN & PF & uni. & TN & (\%) \\
\midrule
LSAFNet & 0.681 & 0.691 & 0.105 & 0.109 & 0.125 & 0.384 & 0.427 & 26.9 & 26.8 & 0.731 & 0.743 & 0.0353 & 0.0076 & 0.0054 & 2.0 \\
FoBa & 0.665 & 0.673 & 0.091 & 0.104 & 0.119 & 0.356 & 0.433 & 25.1 & 24.8 & 0.698 & 0.714 & 0.0404 & 0.0065 & 0.0046 & 2.6 \\
ChangeMask & 0.672 & 0.697 & 0.081 & 0.117 & 0.117 & 0.295 & 0.387 & 51.4 & 35.2 & 0.741 & 0.777 & 0.0585 & 0.0143 & 0.0143 & 0.0 \\
CPGNet & 0.654 & 0.673 & 0.080 & 0.108 & 0.111 & 0.313 & 0.418 & 49.2 & 36.8 & 0.693 & 0.737 & 0.0629 & 0.0149 & 0.0143 & 0.5 \\
SSCD-L & 0.670 & 0.692 & 0.080 & 0.111 & 0.115 & 0.300 & 0.402 & 45.7 & 34.8 & 0.713 & 0.750 & 0.0625 & 0.0165 & 0.0157 & 0.9 \\
BGSNet & 0.642 & 0.639 & 0.078 & 0.086 & 0.087 & 0.305 & 0.339 & 46.4 & 58.9 & 0.679 & 0.706 & 0.0559 & 0.0261 & 0.0255 & 0.6 \\
ChangeMamba & 0.663 & 0.665 & 0.074 & 0.103 & 0.112 & 0.270 & 0.376 & 57.8 & 37.3 & 0.697 & 0.729 & 0.0624 & 0.0161 & 0.0137 & 1.4 \\
BiSRNet & 0.641 & 0.658 & 0.073 & 0.103 & 0.107 & 0.279 & 0.396 & 61.7 & 38.2 & 0.681 & 0.714 & 0.0713 & 0.0187 & 0.0174 & 1.3 \\
TED & 0.659 & 0.683 & 0.071 & 0.096 & 0.098 & 0.272 & 0.367 & 57.6 & 41.7 & 0.706 & 0.744 & 0.0609 & 0.0167 & 0.0162 & 0.6 \\
HRSCD-str.4 & 0.371 & 0.378 & 0.065 & 0.095 & 0.095 & 0.252 & 0.352 & 62.6 & 42.2 & 0.385 & 0.405 & 0.0666 & 0.0115 & 0.0115 & 0.0 \\
HGINet & 0.580 & 0.581 & 0.059 & 0.094 & 0.095 & 0.252 & 0.377 & 74.3 & 49.7 & 0.611 & 0.667 & 0.0887 & 0.0195 & 0.0188 & 1.1 \\
HRSCD-str.3 & 0.359 & 0.360 & 0.058 & 0.086 & 0.086 & 0.216 & 0.329 & 81.0 & 45.3 & 0.372 & 0.394 & 0.0786 & 0.0120 & 0.0120 & 0.0 \\
EGMSNet & 0.597 & 0.617 & 0.055 & 0.093 & 0.096 & 0.229 & 0.389 & 66.1 & 27.7 & 0.636 & 0.676 & 0.0691 & 0.0100 & 0.0092 & 2.3 \\
GSTM-SCD & 0.519 & 0.505 & 0.054 & 0.110 & 0.107 & 0.211 & 0.395 & 77.2 & 28.9 & 0.573 & 0.595 & 0.0762 & 0.0053 & 0.0053 & 1.4 \\
CdSC & 0.507 & 0.523 & 0.054 & 0.064 & 0.064 & 0.218 & 0.247 & 81.9 & 69.3 & 0.548 & 0.594 & 0.0869 & 0.0235 & 0.0235 & 0.0 \\
Change3D & 0.432 & 0.440 & 0.035 & 0.055 & 0.066 & 0.122 & 0.270 & 84.4 & 31.1 & 0.460 & 0.465 & 0.0845 & 0.0139 & 0.0046 & 7.7 \\
\midrule
TRACE-Net (three-seed means) & 0.721 & 0.734 & 0.108 & 0.148 & 0.152 & 0.392 & 0.484 & 20.3 & 19.2 & 0.778 & 0.789 & 0.0308 & 0.0057 & 0.0055 & 1.2 \\
\bottomrule
\end{tabular*}
\end{table*}

\begin{table*}[!htbp]
\centering
\caption{Cross-mine association between area increments and production rates for the ground truth, the per-frame maps (PF) and TRACE-Net (TN), with every site mapped by a network not trained on it (three-seed means). Retention is shown for ground-truth $|\rprod|>0.3$ (n/a otherwise); (z): standardised within commodity groups.}
\label{tab:retention}
\footnotesize
\setlength{\tabcolsep}{3.5pt}
\begin{tabular*}{\linewidth}{@{\extracolsep{\fill}}llrrrrrrrrr@{}}
\toprule
 & & & \multicolumn{2}{c}{ground truth} & \multicolumn{2}{c}{predicted \rprod{}} &
 \multicolumn{2}{c}{predicted \rpart{}} & \multicolumn{2}{c}{retention} \\
\cmidrule(lr){4-5}\cmidrule(lr){6-7}\cmidrule(lr){8-9}\cmidrule(lr){10-11}
Production & Class & $n$ & \rprod{} & \rpart{} & PF & TN & PF & TN & PF & TN \\
\midrule
ore processed & mining & 11 & +0.41 & $-0.26$ & +0.54 & +0.53 & $-0.13$ & $-0.18$ & +1.30 & +1.28 \\
 & pit & 11 & +0.74 & +0.24 & +0.42 & +0.44 & +0.21 & +0.25 & +0.56 & +0.59 \\
 & rock storage & 11 & +0.20 & $-0.36$ & +0.49 & +0.42 & $-0.22$ & $-0.35$ & n/a & n/a \\
 & tailings & 11 & +0.57 & +0.06 & +0.39 & +0.48 & +0.04 & +0.07 & +0.69 & +0.85 \\
 & infrastructure & 11 & $-0.25$ & $-0.26$ & +0.43 & +0.46 & +0.06 & +0.04 & n/a & n/a \\
 & water & 11 & +0.02 & $-0.12$ & $-0.41$ & $-0.19$ & $-0.71$ & $-0.60$ & n/a & n/a \\
\midrule
ore mined & mining & 12 & +0.68 & +0.61 & +0.70 & +0.71 & +0.68 & +0.63 & +1.04 & +1.05 \\
 & pit & 12 & +0.74 & +0.88 & +0.37 & +0.54 & +0.54 & +0.52 & +0.50 & +0.73 \\
 & rock storage & 12 & +0.49 & +0.36 & +0.57 & +0.55 & +0.46 & +0.44 & +1.16 & +1.14 \\
 & tailings & 12 & +0.70 & +0.65 & +0.66 & +0.70 & +0.69 & +0.64 & +0.94 & +1.00 \\
 & infrastructure & 12 & $-0.30$ & $-0.22$ & +0.58 & +0.52 & +0.59 & +0.42 & $-1.92$ & $-1.73$ \\
 & water & 12 & $-0.22$ & $-0.46$ & +0.12 & +0.09 & +0.41 & +0.01 & n/a & n/a \\
\midrule
total material mined & mining & 6 & +0.52 & $-0.18$ & +0.57 & +0.57 & +0.23 & +0.18 & +1.10 & +1.09 \\
 & pit & 6 & +0.78 & $-0.26$ & +0.57 & +0.71 & +0.78 & +0.56 & +0.72 & +0.91 \\
 & rock storage & 6 & +0.54 & $-0.11$ & +0.59 & +0.53 & +0.16 & +0.13 & +1.09 & +0.99 \\
 & tailings & 6 & +0.38 & $-0.35$ & +0.30 & +0.47 & $-0.06$ & $-0.02$ & +0.80 & +1.24 \\
 & infrastructure & 6 & $-0.68$ & $-0.01$ & +0.78 & +0.82 & +0.92 & +0.97 & $-1.15$ & $-1.20$ \\
 & water & 6 & $-0.01$ & +0.59 & +0.51 & +0.66 & +0.93 & +0.95 & n/a & n/a \\
\midrule
main-commodity output (z) & mining & 12 & +0.52 & +0.17 & $-0.31$ & +0.53 & 0.00 & +0.25 & $-0.59$ & +1.02 \\
 & pit & 12 & +0.48 & +0.03 & $-0.37$ & +0.48 & $-0.17$ & +0.16 & $-0.76$ & +0.99 \\
 & rock storage & 12 & +0.52 & +0.20 & $-0.13$ & +0.49 & +0.06 & +0.21 & $-0.26$ & +0.94 \\
 & tailings & 12 & +0.49 & +0.16 & $-0.39$ & +0.49 & $-0.03$ & +0.22 & $-0.79$ & +1.01 \\
 & infrastructure & 12 & +0.22 & $-0.26$ & $-0.63$ & +0.06 & $-0.42$ & $-0.12$ & n/a & n/a \\
 & water & 12 & $-0.42$ & +0.24 & $-0.06$ & 0.00 & $-0.03$ & $-0.01$ & +0.14 & +0.01 \\
\bottomrule
\end{tabular*}
\end{table*}

\begin{table*}[!htbp]
\centering
\caption{Inference variants and settings of TRACE-Net on the test sites (complete method, three-seed means); each row changes one setting. MineNetTrace: endpoint F1, \TSeK{}, mining-area $|\dA|$ err (\si{\hectare}), intermediate F1 and pixel flip rate; EuroMineNet: F1 over all years, share of single changes dated within one year and annual increment fidelity. Entropy-weighted mean: each prediction weighted by one minus its entropy divided by the logarithm of the number of classes. Intermediate acquisitions: scored on the retained acquisitions; dash: not comparable with the full series.}
\label{tab:infer}
\footnotesize
\setlength{\tabcolsep}{3pt}
\begin{tabular*}{\linewidth}{@{\extracolsep{\fill}}lrrrrrrrr@{}}
\toprule
 & \multicolumn{5}{c}{MineNetTrace} & \multicolumn{3}{c}{EuroMineNet} \\
\cmidrule(lr){2-6}\cmidrule(lr){7-9}
Setting & F1 & \TSeK{} & \colhead[r]{$|\dA|$ err\\mining} & \colhead[r]{inter-\\mediate F1} & \colhead[r]{pixel\\flip rate} & \colhead[r]{F1,\\all years} & \colhead[r]{dated within\\one year} & \colhead[r]{annual\\fidelity} \\
\midrule
TRACE-Net (bidirectional, agreement-weighted; $\omega=4$, $w=0.5$) & 0.734 & 0.152 & 37.2 & 0.789 & 0.0055 & 0.881 & 0.618 & +0.21 \\
\multicolumn{9}{@{}l}{\emph{anchored inference}} \\
\quad unidirectional & 0.735 & 0.148 & 38.0 & 0.790 & 0.0057 & 0.880 & 0.616 & +0.17 \\
\quad bidirectional, plain mean & 0.734 & 0.152 & 37.4 & 0.789 & 0.0057 & 0.881 & 0.617 & +0.21 \\
\quad bidirectional, entropy-weighted mean & 0.735 & 0.152 & 37.6 & 0.789 & 0.0058 & 0.880 & 0.614 & +0.19 \\
\multicolumn{9}{@{}l}{\emph{evidence weight $\omega$}} \\
\quad 1 & 0.731 & 0.142 & 40.4 & 0.788 & 0.0048 & 0.880 & 0.622 & +0.24 \\
\quad 2 & 0.733 & 0.149 & 39.2 & 0.789 & 0.0048 & 0.881 & 0.620 & +0.24 \\
\quad 8 & 0.734 & 0.145 & 35.9 & 0.788 & 0.0074 & 0.880 & 0.617 & +0.20 \\
\multicolumn{9}{@{}l}{\emph{change weight $w$}} \\
\quad 0 (trajectory inference) & 0.733 & 0.131 & 54.3 & 0.789 & 0.0067 & 0.881 & 0.592 & +0.19 \\
\quad 0.25 & 0.734 & 0.149 & 40.0 & 0.789 & 0.0054 & 0.881 & 0.609 & +0.22 \\
\quad 1 & 0.733 & 0.149 & 39.6 & 0.789 & 0.0066 & 0.880 & 0.628 & +0.20 \\
\quad 2 & 0.730 & 0.141 & 43.0 & 0.788 & 0.0083 & 0.879 & 0.636 & +0.19 \\
\multicolumn{9}{@{}l}{\emph{change rate of the prior}} \\
\quad $\times$0.25 & 0.734 & 0.151 & 39.7 & 0.789 & 0.0049 & 0.881 & 0.614 & +0.23 \\
\quad $\times$0.5 & 0.734 & 0.152 & 38.4 & 0.789 & 0.0052 & 0.881 & 0.616 & +0.22 \\
\quad $\times$2 & 0.734 & 0.152 & 36.7 & 0.789 & 0.0059 & 0.881 & 0.620 & +0.21 \\
\quad $\times$4 & 0.734 & 0.151 & 36.9 & 0.788 & 0.0063 & 0.880 & 0.622 & +0.21 \\
\multicolumn{9}{@{}l}{\emph{intermediate acquisitions}} \\
\quad every second removed & 0.731 & 0.147 & 39.8 & 0.789 & 0.0046 & 0.880 & -- & -- \\
\quad random half (three draws) & 0.731 & 0.152 & 39.6 & 0.787 & 0.0046 & 0.880 & -- & -- \\
\quad middle one only & 0.726 & 0.143 & 56.5 & 0.778 & 0.0039 & 0.880 & -- & -- \\
\bottomrule
\end{tabular*}
\end{table*}

\begin{table}[!htbp]
\centering
\caption{Single-factor ablations of TRACE-Net on the test sites (complete method, three-seed means). Each row changes one setting; the setting of TRACE-Net is given in the group heading. M: million parameters; F1: endpoint F1; $\Delta$: change in \TSeK{} relative to TRACE-Net; robust: standardised with the per-band median and interquartile range.}
\label{tab:ablation_main}
\footnotesize
\setlength{\tabcolsep}{3pt}
\begin{tabular*}{\linewidth}{@{}>{\raggedright\arraybackslash\hangindent=1em\hangafter=1}p{0.56\linewidth}@{\extracolsep{\fill}}rrr@{}}
\toprule
Setting & F1 & \TSeK{} & $\Delta$\TSeK{} \\
\midrule
TRACE-Net (SegFormer-B2, 26.1\,M) & 0.734 & 0.152 & -- \\
\multicolumn{4}{@{}l}{\emph{Backbone} (TRACE-Net: SegFormer-B2, 26.1\,M)} \\
\quad U-Net-R34, 25.5\,M & 0.706 & 0.108 & $-0.044$ \\
\quad DeepLabv3+-R50, 43.8\,M & 0.708 & 0.126 & $-0.026$ \\
\quad UPerNet-ConvNeXt-T, 39.7\,M & 0.732 & 0.148 & $-0.004$ \\
\multicolumn{4}{@{}l}{\emph{Change branch} (TRACE-Net: lightweight head on the decoder features)} \\
\quad SCD-style change decoder, 27.0\,M & 0.728 & 0.147 & $-0.005$ \\
\multicolumn{4}{@{}l}{\emph{Initialisation} (TRACE-Net: ImageNet encoder)} \\
\quad random encoder & 0.415 & 0.059 & $-0.093$ \\
\multicolumn{4}{@{}l}{\emph{Normalisation} (TRACE-Net: global, training sites)} \\
\quad fixed reflectance scaling & 0.731 & 0.149 & $-0.003$ \\
\quad global statistics of all sites & 0.731 & 0.148 & $-0.004$ \\
\quad per-site robust & 0.715 & 0.151 & $-0.001$ \\
\quad per-frame robust & 0.707 & 0.149 & $-0.003$ \\
\multicolumn{4}{@{}l}{\emph{Class-weight exponent} (TRACE-Net: 0.5)} \\
\quad 0 (unweighted) & 0.728 & 0.140 & $-0.011$ \\
\quad 1.0 (full inverse frequency) & 0.702 & 0.142 & $-0.009$ \\
\multicolumn{4}{@{}l}{\emph{Tile sampling} (TRACE-Net: weights 8:2:1)} \\
\quad change tiles only & 0.719 & 0.148 & $-0.004$ \\
\quad natural frequency & 0.728 & 0.139 & $-0.013$ \\
\bottomrule
\end{tabular*}
\end{table}

\FloatBarrier
\section{Seed standard deviations}
\label{app:sd}
\setcounter{table}{0}

Tables~\ref{tab:mapping} and~\ref{tab:design} report three-seed means. Tables~\ref{tab:mapping_sd} and~\ref{tab:design_sd} give the seed standard deviations of the same entries, with one decimal more than the means.

\begin{table*}[!htbp]
\centering
\caption{Seed standard deviations for Table~\ref{tab:mapping}, in the same layout.}
\label{tab:mapping_sd}
\footnotesize
\setlength{\tabcolsep}{1.5pt}
\begin{tabular*}{\linewidth}{@{\extracolsep{\fill}}lrrrrrrrrr@{\hspace{7pt}}rrrrr@{}}
\toprule
 & \multicolumn{9}{c}{MineNetTrace} & \multicolumn{5}{c}{EuroMineNet} \\
\cmidrule(lr){2-10}\cmidrule(lr){11-15}
 & & & \multicolumn{2}{c}{$|\dA|$ err} & & & & & & & & \multicolumn{2}{c}{dated} & \\
\cmidrule(lr){4-5}\cmidrule(lr){13-14}
 & F1 & \TSeK{} & mining & \colhead[r]{class\\mean} & \colhead[r]{\rinc{}\\class\\mean} & \colhead[r]{inter-\\mediate\\F1} & \colhead[r]{pixel\\flip\\rate} & \colhead[r]{change\\recall} & \colhead[r]{$E$,\\class\\mean} & \colhead[r]{F1,\\all\\years} & \colhead[r]{acc.,\\changed} & \colhead[r]{correct\\year} & \colhead[r]{within\\one\\year} & \colhead[r]{annual\\fidelity} \\
\midrule
\multicolumn{15}{@{}l}{\emph{training on the endpoint masks}} \\
\quad per-frame maps & 0.0018 & 0.0014 & 3.11 & 1.90 & 0.0578 & 0.0005 & 0.00045 & 0.0050 & 0.0143 & 0.0020 & 0.0027 & 0.0033 & 0.0065 & 0.011 \\
\quad first-endpoint persistence & 0.0018 & 0.0014 & 3.11 & 1.90 & 0.0578 & 0.0005 & 0.00048 & 0.0050 & 0.0000 & 0.0028 & 0.0012 & 0.0006 & 0.0013 & 0.011 \\
\quad nearest endpoint & 0.0018 & 0.0014 & 3.11 & 1.90 & 0.0578 & 0.0005 & 0.00048 & 0.0050 & 0.0000 & 0.0032 & 0.0025 & 0.0006 & 0.0012 & 0.096 \\
\quad three-frame majority filter & 0.0018 & 0.0014 & 3.11 & 1.90 & 0.0578 & 0.0002 & 0.00049 & 0.0050 & 0.0241 & 0.0018 & 0.0020 & 0.0019 & 0.0081 & 0.023 \\
\quad break detection & 0.0026 & 0.0015 & 3.18 & 0.75 & 0.0533 & 0.0019 & 0.00027 & 0.0057 & 0.0167 & 0.0020 & 0.0013 & 0.0044 & 0.0065 & 0.010 \\
\quad Markov-chain inference & 0.0039 & 0.0006 & 0.41 & 0.72 & 0.0812 & 0.0022 & 0.00022 & 0.0040 & 0.0251 & 0.0019 & 0.0012 & 0.0029 & 0.0095 & 0.014 \\
\quad trajectory inference & 0.0040 & 0.0005 & 0.58 & 0.85 & 0.0774 & 0.0022 & 0.00022 & 0.0040 & 0.0155 & 0.0019 & 0.0010 & 0.0016 & 0.0077 & 0.024 \\
\quad + change evidence (TRACE-Net) & 0.0039 & 0.0028 & 1.40 & 1.48 & 0.0190 & 0.0023 & 0.00035 & 0.0107 & 0.0313 & 0.0021 & 0.0011 & 0.0036 & 0.0065 & 0.024 \\
\multicolumn{15}{@{}l}{\emph{training on every annual mask (EuroMineNet only, reference)}} \\
\quad per-frame maps & -- & -- & -- & -- & -- & -- & -- & -- & -- & 0.0006 & 0.0038 & 0.0135 & 0.0136 & 0.050 \\
\quad trajectory inference + change evidence & -- & -- & -- & -- & -- & -- & -- & -- & -- & 0.0002 & 0.0019 & 0.0037 & 0.0055 & 0.016 \\
\bottomrule
\end{tabular*}
\end{table*}

\begin{table*}[!htbp]
\centering
\caption{Seed standard deviations for Table~\ref{tab:design}, in the same layout.}
\label{tab:design_sd}
\footnotesize
\setlength{\tabcolsep}{4pt}
\begin{tabular*}{\linewidth}{@{\extracolsep{\fill}}lrrrrrrrrr@{}}
\toprule
 & \multicolumn{2}{c}{per-frame maps} & TI & \multicolumn{6}{c}{complete method} \\
\cmidrule(lr){2-3}\cmidrule(lr){4-4}\cmidrule(lr){5-10}
Variant & F1 & \TSeK{} & \TSeK{} & F1 & \TSeK{} & $F_{\mathrm{scd}}$ & \colhead[r]{\rinc{}\\mining} & \colhead[r]{\rinc{}\\class mean} & \colhead[r]{$|\dA|$ err\\class mean} \\
\midrule
no coupling & 0.0050 & 0.0037 & 0.0044 & 0.0059 & 0.0044 & 0.0144 & 0.0167 & 0.0742 & 3.67 \\
change branch only & 0.0053 & 0.0017 & 0.0054 & 0.0125 & 0.0031 & 0.0024 & 0.0254 & 0.0326 & 0.44 \\
fusion only & 0.0082 & 0.0042 & 0.0065 & 0.0092 & 0.0065 & 0.0124 & 0.0284 & 0.0798 & 0.48 \\
fusion + change branch (TRACE-Net) & 0.0018 & 0.0014 & 0.0005 & 0.0039 & 0.0028 & 0.0004 & 0.0110 & 0.0190 & 1.48 \\
\bottomrule
\end{tabular*}
\end{table*}

\end{document}